\documentclass[10pt,twocolumn,letterpaper]{article}
\usepackage[pagenumbers]{cvpr}

\usepackage{booktabs}
\usepackage{multirow}
\usepackage{array}
\usepackage{graphicx}
\usepackage{caption}
\usepackage{adjustbox}
\usepackage{makecell}
\usepackage[table]{xcolor}
\usepackage[normalem]{ulem}
\usepackage{pifont}
\usepackage{amsmath}
\usepackage{ulem}
\usepackage[accsupp]{axessibility}  
\definecolor{cvprblue}{rgb}{0.21,0.49,0.74}
\usepackage[pagebackref,breaklinks,colorlinks,allcolors=cvprblue]{hyperref}

\def\paperID{35277}  
\def\confName{CVPR}
\def\confYear{2026}

\title{EW-DETR: \underline{E}volving \underline{W}orld Object Detection via \\ Incremental Low-Rank \underline{DE}tection \underline{TR}ansformer}

\author{
Munish Monga$^1$ \quad Vishal Chudasama$^1$ \quad Pankaj Wasnik$^1$ \quad C.V. Jawahar$^2$ \\ 
$^1$Sony Research India \quad $^2$IIIT Hyderabad\\ 
{\tt\small \{munish.monga, vishal.chudasama1, pankaj.wasnik\}@sony.com, jawahar@iiit.ac.in} 
}
\definecolor{lightyellow}{HTML}{FFFF00}
\definecolor{lightblue}{HTML}{BDD7EE}   
\definecolor{lightred}{HTML}{F8CBAD}    
\definecolor{lightgreen}{HTML}{C5E0B4}  
\definecolor{lightpurple}{HTML}{F3EAFD} 
\definecolor{lightgrey}{HTML}{D9D9D9}
\definecolor{darkred}{HTML}{8B0000}
\definecolor{darkgreen}{HTML}{006400}
\definecolor{verylightyellow}{HTML}{FFFFCC}
\definecolor{verylightblue}{HTML}{E6F2FF}
\definecolor{verylightorange}{HTML}{FFEBD6}
\definecolor{lightlavender}{HTML}{F0E6FF}
\definecolor{lightgrayblue}{HTML}{E8EEF7}
\definecolor{vocblue}{RGB}{30,90,200}
\definecolor{cliporange}{RGB}{230,140,30}
\definecolor{waterteal}{RGB}{20,160,160}
\definecolor{comicmagenta}{RGB}{200,40,150}
\definecolor{unkred}{RGB}{200,0,0}

\newcommand{\cmark}{\ding{51}}

\newcommand{\deltaa}[1]{\ifnum #1>0 {\color{red}\textuparrow\ #1} \else \ifnum #1<0 {\color{blue}\textdownarrow\ #1} \else #1\fi\fi}

\begin{document}
\twocolumn[{
    \renewcommand\twocolumn[1][]{#1} 
    \vspace{-8mm}
    \maketitle 
    \vspace{-12mm}
    \begin{center}
        \centering
        \includegraphics[width=0.85\linewidth]{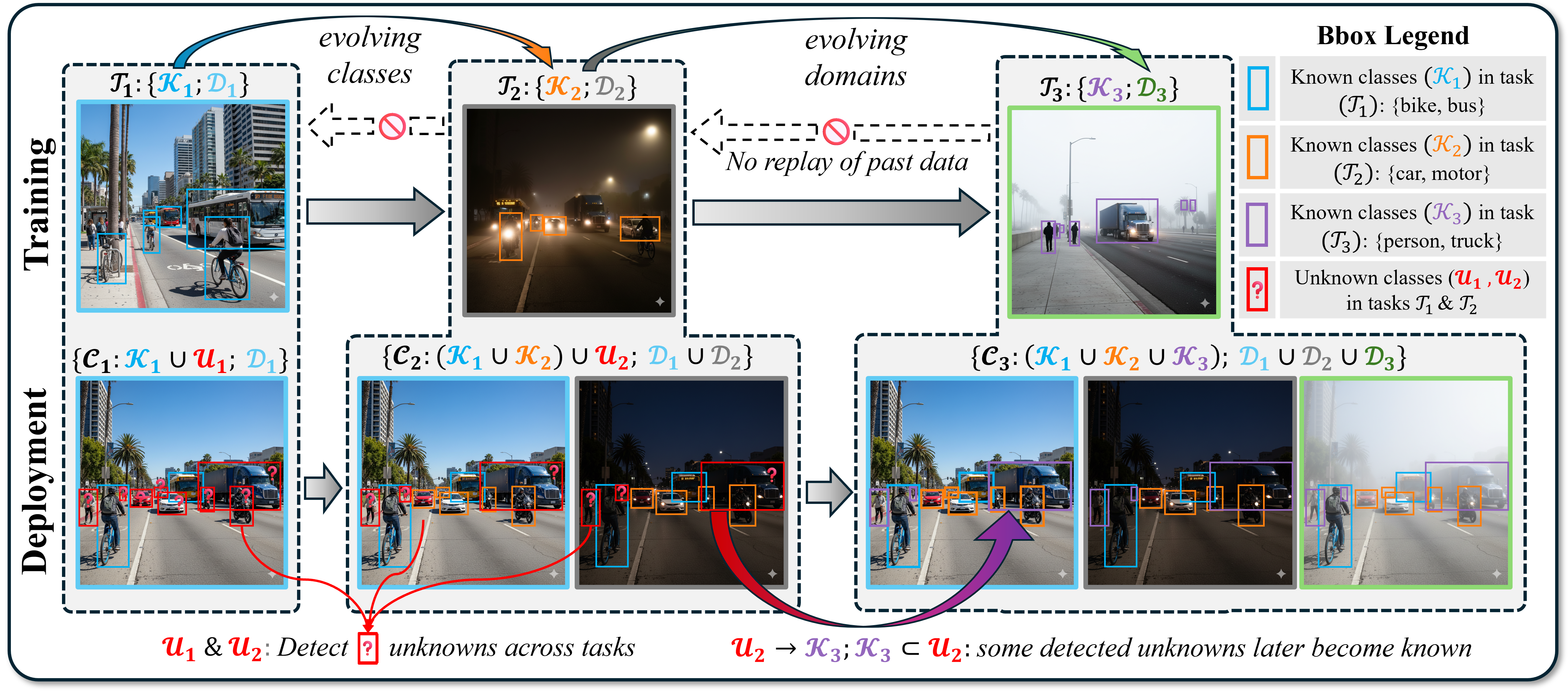}
        \vspace{-2mm}
        \captionsetup{hypcap=false}
        \captionof{figure}{\textbf{Overview of Evolving World Object Detection (EWOD).} \textbf{Top (Training)}: Sequential tasks introduce disjoint classes \(\mathcal{K}_t\) across shifting domains \(\mathcal{D}_t\) (Day \(\to\) Night \(\to\) Fog). \textbf{Bottom (Deployment)}: The detector must
        (i) identify unseen objects \(\mathcal{U}_t\) without supervision; 
        (ii) retain knowledge of all prior classes across all domains; 
        (iii) incrementally learn former unknowns when their labels are revealed later (e.g., \(\mathcal{U}_2 \to \mathcal{K}_3\)) and 
        (iv) achieve all these objectives without revisiting any previous data. Best viewed in colour with zoom\protect\footnotemark[1].}
    \label{fig:teaser}
    \end{center}
}]
\begin{abstract}
Real-world object detection must operate in evolving environments where new classes emerge, domains shift, and unseen objects must be identified as \textit{``unknown''}—all without accessing prior data. We introduce Evolving World Object Detection (EWOD), a paradigm coupling incremental learning, domain adaptation, and unknown detection under exemplar-free constraints. To tackle EWOD, we propose EW-DETR framework that augments DETR-based detectors with three synergistic modules: Incremental LoRA Adapters for exemplar-free incremental learning under evolving domains; a Query-Norm Objectness Adapter that decouples objectness-aware features from DETR decoder queries; and Entropy-Aware Unknown Mixing for calibrated unknown detection. This framework generalises across DETR-based detectors, enabling state-of-the-art RF-DETR to operate effectively in evolving-world settings. We also introduce FOGS (Forgetting, Openness, Generalisation Score) to holistically evaluate performance across these dimensions. Extensive experiments on Pascal Series and Diverse Weather benchmarks show EW-DETR outperforms other methods, improving FOGS by \textbf{57.24}\%.
\end{abstract}

\section{Introduction}
\label{sec:intro}
Object detection has witnessed remarkable progress in recent years, yet most state-of-the-art detectors remain confined to closed-world assumptions~\cite{RF-DETR, DINO, DETR, D-DETR}. However, real-world deployment scenarios demand fundamentally different capabilities. Consider an autonomous vehicle: it must continuously identify new object types (construction equipment, novel vehicle models), adapt to diverse environmental conditions (day to night, clear to foggy weather), and critically, recognise unseen objects as \textit{``unknown''} to avoid catastrophic failures. Similarly, warehouse robots must handle evolving product inventories while adapting to varying lighting and seasonal conditions, all without revisiting past training data. These scenarios exemplify the need for object detectors that can operate in an \textbf{Evolving World}.

\begin{figure}[t!]
\begin{center}
\centerline{\includegraphics[width=\columnwidth]{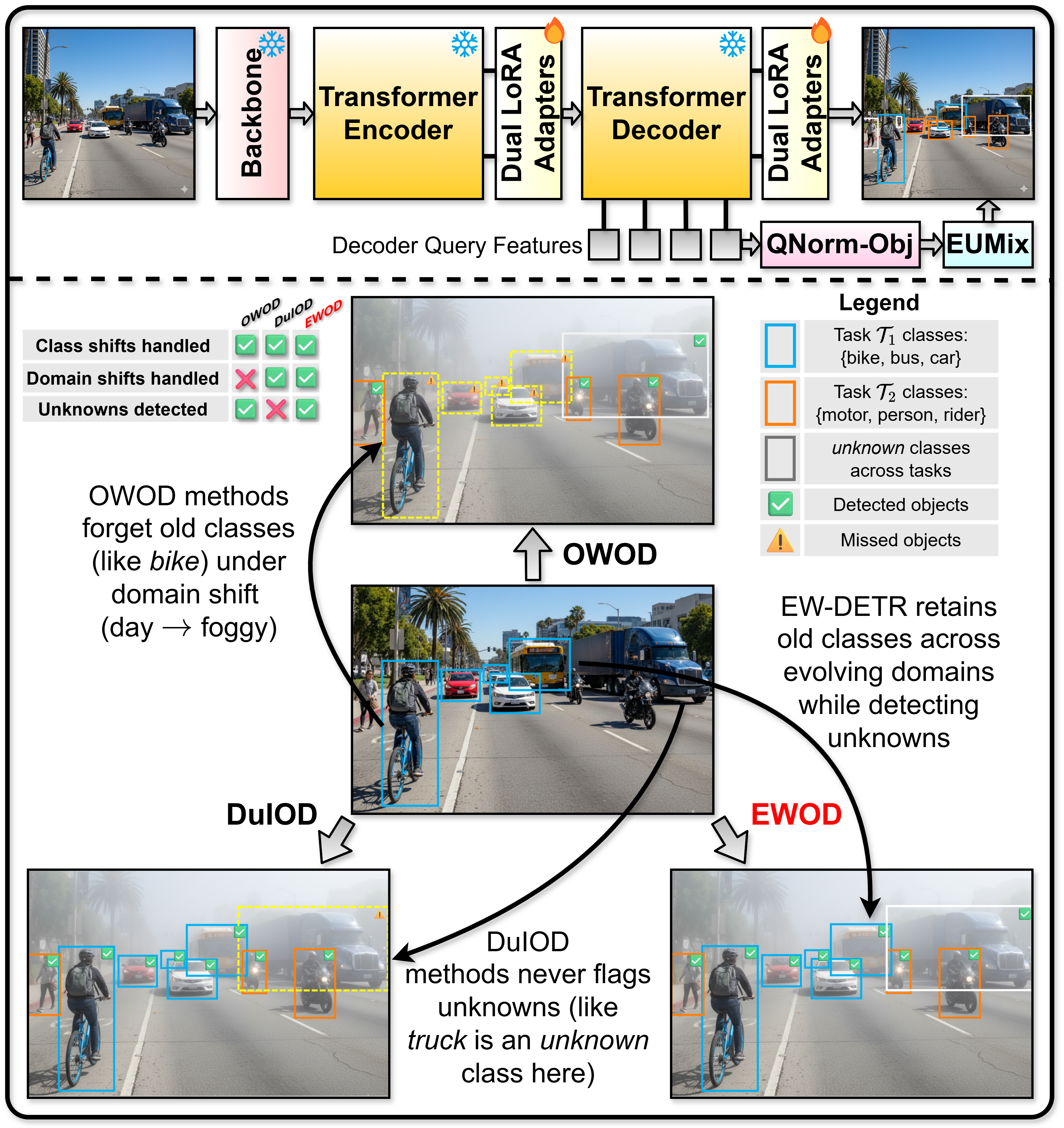}}
\vspace{-1em}
\caption{\textbf{Top Row:} gives an overview of the EW-DETR framework, which augments the standard DETR pipeline with three key modules: \emph{Incremental LoRA Adapters} for exemplar-free incremental learning under evolving domains, \emph{Query-Norm Objectness Adapter} for decoupled objectness-aware features, and \emph{Entropy-Aware Unknown Mixing} for calibrated unknown detection. \textbf{Bottom Row:} illustrates a quick qualitative comparison of EWOD with other OD approaches. Best viewed in colour with zoom\protect\footnotemark[1].}
\vspace{-10.5mm}
\label{fig:intro}
\end{center}
\end{figure}

We introduce \textbf{Evolving World Object Detection (EWOD)}, a novel paradigm that reflects how object detectors must operate in real, dynamic, and continuously evolving environments. As illustrated in Figure~\ref{fig:teaser}, EWOD couples three evolving dimensions: new object classes emerge across tasks, visual domains shift significantly (day \(\to\) night \(\to\) fog), and truly novel objects must be identified as \textit{``unknown''} without explicit supervision. More importantly, detectors must handle all of this without storing or revisiting any previous training data. 

Unlike EWOD, existing paradigms address only subsets of these challenges. Open-World Object Detection (OWOD)~\cite{ORE,OW-DETR,CAT,PROB,OWOBJ} focuses on detecting unknowns and incrementally learning new classes, but critically assumes a single static domain and relies heavily on exemplar replay—storing representative samples from previous tasks to mitigate catastrophic forgetting. This limits scalability when data storage is limited or when privacy regulations restrict data retention, such as in healthcare systems~\cite{verma2023privacy}. Other recent approaches like Domain Incremental Object Detection (DIOD)~\cite{LDB} and Dual Incremental Object Detection (DuIOD)~\cite{DuET} operate under closed-world assumptions where all foreground objects are assumed to belong to known categories. As shown in Figure~\ref{fig:intro}, when deployed in evolving-world settings, such detectors fail catastrophically: they either misclassify unknown objects into known categories (leading to overconfident, incorrect predictions) or absorb them into the background class (causing missed detections of potentially critical novel objects). 

EWOD introduces fundamentally harder challenges than standard OWOD, where domain consistency provides a stable feature space for unknown detection, or DuIOD, where a closed label space eliminates the need for calibrated unknown modelling. Furthermore, EWOD faces a severe data imbalance across tasks, as different domains and class distributions yield vastly different sample sizes per task. Moreover, EWOD restricts access to previous data and eliminates replay-based rehearsal, which is used by OWOD methods~\cite{ORE,OW-DETR,CAT,PROB,OWOBJ}, requiring alternative approaches.

To address these challenges, we propose \textbf{EW-DETR}, an exemplar-free framework that augments DETR-based detectors~\cite{DETR,D-DETR,RF-DETR} with three synergistic modules (Figure~\ref{fig:intro}). First, \emph{Incremental LoRA Adapters} (Section~\ref{sec:incre_lora}) employ a dual-adapter architecture: an aggregate adapter that accumulates compressed knowledge from all previous tasks and a task-specific adapter that captures current task updates. Through data-aware merging guided by per-task sample ratios and low-rank projection via truncated SVD, we achieve stable knowledge consolidation without storing any exemplars while explicitly addressing data imbalance. Second, \emph{Query-Norm Objectness Adapter} (Section~\ref{sec:query_norm_objectness}) decouples query semantics from magnitude by normalizing decoder features, yielding domain-invariant class-agnostic representations that enable robust unknown detection even under severe domain shifts—crucially, without requiring any auxiliary supervision or additional losses, unlike \cite{ORTH, ZiRA}. Third, \emph{Entropy-Aware Unknown Mixing} (Section~\ref{sec:unknown_mix}) calibrates unknown predictions by combining classification uncertainty with objectness evidence, ensuring that high-objectness, high-uncertainty queries are correctly identified as unknowns rather than spuriously absorbed into known classes or background. 

We highlight our contributions as:
\begin{itemize}
    \item To the best of our knowledge, EW-DETR is the first framework for the proposed EWOD paradigm, simultaneously tackling incrementally evolving classes, generalising across shifting domains, and detecting unknown objects, all under strict exemplar-free constraints.
    \item We propose a novel incremental dual LoRA adapter mechanism with data-aware merging that effectively mitigates catastrophic forgetting under joint class-domain evolution without storing any previous data.
    \item We introduce Query-Norm Objectness Adapter that operates on class-agnostic decoder queries of DETR-based detectors with an Entropy-Aware Unknown Mixing module to enable robust unknown detection without any auxiliary losses, helping current SOTA RF-DETR~\cite{RF-DETR} to operate in evolving-world settings.
    \item We introduce FOGS (Forgetting, Openness, Generalisation Score), a comprehensive metric that holistically evaluates detector performance across the three critical dimensions of EWOD: retention of past knowledge, detection of unknowns, and generalisation across domains.
\end{itemize}

\footnotetext[1]{Images are generated using Gemini 2.5 Pro for illustration purposes.}

\section{Related Works}
\label{sec:related_works}
\paragraph{Open-World Object Detection.}
Open-world object detection (OWOD) extends traditional closed-set detection to identify novel objects as \emph{``unknown''} while incrementally learning their labels when revealed. ORE~\cite{ORE} pioneers this setting by introducing an energy-based unknown identifier and contrastive clustering to discover unknowns. OW-DETR~\cite{OW-DETR} adapts DETR with multi-scale context encoding and pseudo-labelling of high-objectness proposals for unknown detection. CAT~\cite{CAT} employs a cascade architecture with separate localisation and identification stages to decouple unknown discovery from classification. PROB~\cite{PROB} models objectness probabilistically using mixture models to better separate known and unknown distributions. ORTH~\cite{ORTH} subsequently explored orthogonalisation techniques to tackle OWOD.  OWOBJ~\cite{OWOBJ} is a recent method that unifies novel object detection through objectness modelling that generalises across both known and unknown categories. However, these methods assume a single static domain and rely on exemplar replay, failing to address the joint challenges of domain shift and class evolution that real-world deployments face. EWOD requires detectors to simultaneously handle evolving visual domains while maintaining open-world capabilities exemplar-free, which EW-DETR achieves through parameter-efficient dual LoRA adapters and entropy-aware unknown calibration without storing any data.
\vspace{-4mm}
\paragraph{Class (and Domain) Incremental Object Detection.}
Class-incremental object detection (CIOD) methods \cite{pang2019libra, Faster_ILOD, ABR_IOD, cermelli2022modeling, shmelkov2017incremental, SID, RILOD, ERD} addresses the learning of new object categories sequentially while preventing catastrophic forgetting of previously learned ones. However, most approaches rely heavily on exemplar replay and knowledge distillation to maintain performance on old classes \cite{zhong2025replay, Rodeo, SDDGR, CL-DETR}. ORE~\cite{ORE} extends CIOD to the open-world setting but still depends on stored exemplars. Unlike CIOD, Domain-Incremental Object Detection (DIOD) methods \cite{LDB, S-Prompts, DISC, PINA} focus on adapting a detector to a sequence of shifting domains. LDB~\cite{LDB} is a non-exemplar DIOD method that freezes a base detector and learns domain-specific bias terms for each new domain. However, DIOD methods still operate under a closed-set label space where all foreground categories are known a priori. DuET~\cite{DuET} introduced Dual Incremental Object Detection (DuIOD) to handle both class and domain increments through exemplar-free task arithmetic, decomposing model updates into task-generic and task-specific components. However, these approaches largely remain closed-world: unlabeled foreground is treated as background, and no explicit unknown category is modelled. In contrast, EWOD couples domain evolution with a growing label space and explicit unknown detection. EW-DETR addresses these joint challenges by using incremental LoRA adapters to specialise to each new domain, and task-agnostic objectness-based unknown modelling to ensure stability under shifting conditions.

\section{Method}
\label{sec:method}

\begin{figure*}[ht]
\begin{center}
\centerline{\includegraphics[width=0.94\textwidth]{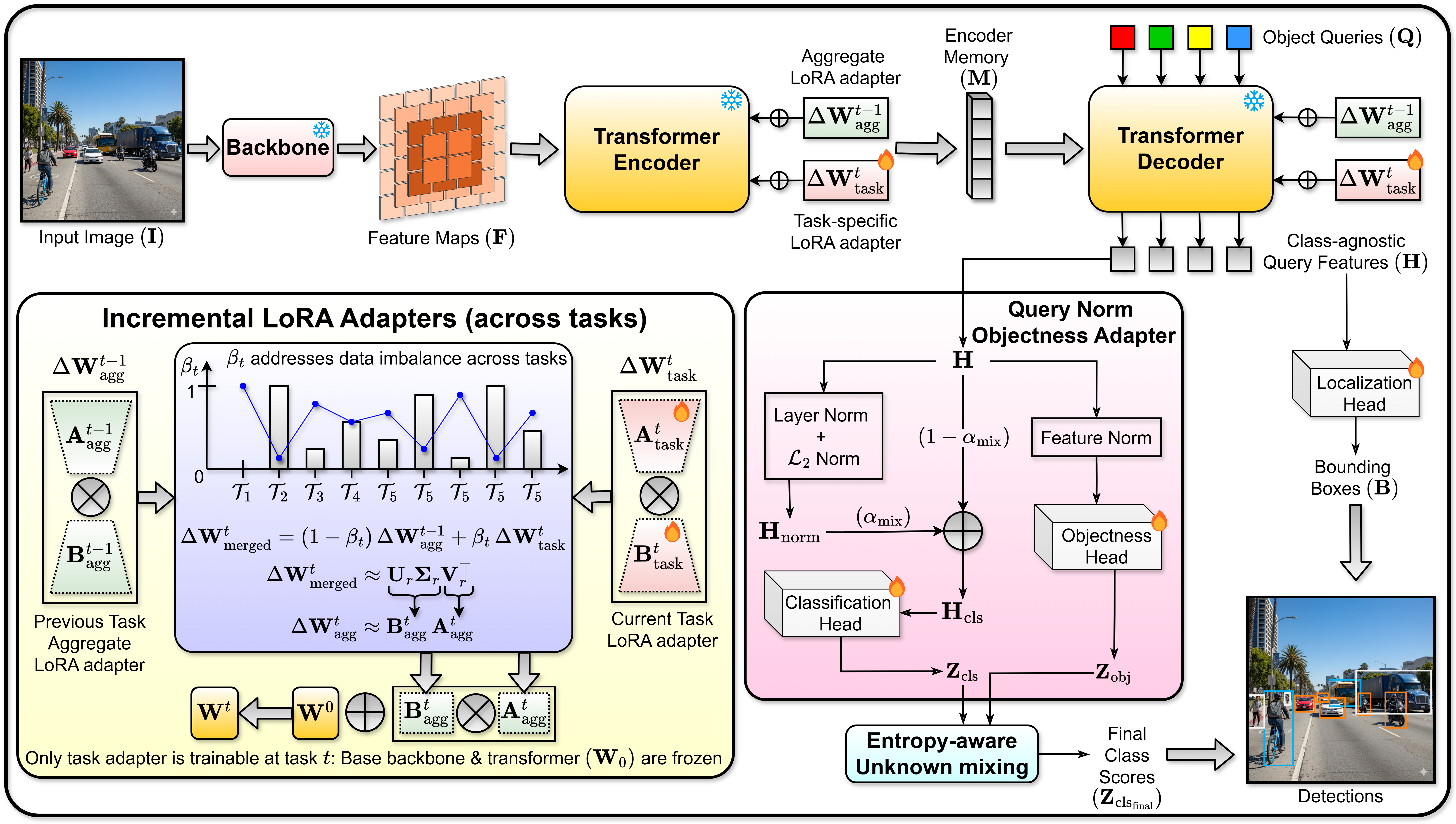}}
\vspace{-3mm}
\caption{
\textbf{EW-DETR framework for Evolving-World Object Detection.} An input image is processed by a frozen backbone and transformer encoder–decoder equipped with aggregate \(\big(\Delta \mathbf{W}^{t-1}_{\text{agg}}\big)\) and task-specific \(\big(\Delta \mathbf{W}^{t}_{\text{task}}\big)\) LoRA adapters, yielding class-agnostic query features. These queries are reparameterised by the Query-Norm Objectness Adapter and passed through a classification head and an objectness head, whose outputs are combined by Entropy-aware Unknown Mixing module to produce calibrated class scores, while a localisation head predicts bounding boxes to form final detections across tasks\protect\footnotemark[1].}
\vspace{-12mm}
\label{fig:main_arch}
\end{center}
\end{figure*}

\subsection{Problem Formulation}
\label{sec:problem_formulation}
In Evolving World Object Detection (EWOD), we consider a sequence of tasks \(\mathcal{T} = \{\mathcal{T}_1, \mathcal{T}_2, \ldots, \mathcal{T}_T\}\), where each task \(\mathcal{T}_t\) provides a dataset \(\mathcal{X}_t \sim P(X_t)\) and introduces a disjoint set of object classes \(\mathcal{K}_t\) from a new domain \(\mathcal{D}_t\) (see Figure~\ref{fig:teaser}). During training on \(\mathcal{T}_t\), annotations are provided only for the known classes \(\mathcal{K}_t\) of the current task; any instances of previously learned classes \(\mathcal{K}_{t-1}\) or truly novel classes \((\mathcal{U}_t)\) that appear in \(\mathcal{X}_t\) remain unlabeled, as in \cite{ORE,PROB}. Moreover, for \(t \geq 2\), the classes are disjoint across tasks \((\mathcal{K}_t \cap \mathcal{K}_{1:t-1} = \varnothing)\), and the image distributions shift between tasks, i.e., for \(t \neq t'\), \(P(X_t) \neq P(X_{t'})\), as in \cite{DuET}. During inference, both the category space and domains evolve incrementally across tasks: 
\begin{equation}
    \mathcal{C}_t = (\mathcal{K}_1 \cup \ldots \cup \mathcal{K}_t) \cup \mathcal{U}_t, \quad
\mathcal{D}_t = (\mathcal{D}_1 \cup \ldots \cup \mathcal{D}_t).
\end{equation}
Hence, EWOD generalizes beyond prior works in OWOD \cite{ORE, OWOBJ} and DuIOD \cite{DuET} and requires object detectors to simultaneously: (i) detect all known classes \(\mathcal{K}^t\) across all seen domains \(\{\mathcal{D}_1, \ldots, \mathcal{D}_t\}\); (ii) detect all unseen objects as \textit{“unknown’’} without explicit unknown supervision; (iii) incrementally learn a subset of unknowns as knowns when their labels are revealed in later tasks \((\mathcal{U}_t \to \mathcal{K}_{t+1}; \mathcal{K}_{t+1} \subseteq \mathcal{U}_t)\); and (iv) achieve these objectives in an exemplar-free manner without storing any previous data \(\{\mathcal{X}_1, \ldots, \mathcal{X}_{t-1}\}\).

\subsection{Overall Architecture}
\label{sec:arch_overview}

EW-DETR framework builds upon DETR-based object detectors~\cite{DETR, D-DETR, RF-DETR} while introducing three key modules to tackle the challenges of EWOD effectively: Incremental LoRA Adapters (Section~\ref{sec:incre_lora}), Query-Norm Objectness Adapter (Section~\ref{sec:query_norm_objectness}), and Entropy-Aware Unknown Mixing (Section~\ref{sec:unknown_mix}). It enables the state-of-the-art DETR-based object detector RF-DETR~\cite{RF-DETR} to operate in the evolving world setting while also generalising to the traditional Deformable DETR detector~\cite{D-DETR} (see Table~\ref{tab:E1_extended}). 

\noindent As shown in Figure~\ref{fig:main_arch}, the EW-DETR framework augments the generic DETR pipeline with three modular components. First, \emph{Incremental LoRA adapters} are attached to all transformer encoder and decoder linear layers, providing a compact memory of past tasks, mitigating catastrophic forgetting without storing any data from previous tasks. Second, a \emph{Query-Norm Objectness Adapter} reparameterises the decoder output features to yield a decoupled representation that helps in \textit{``unknown''} detection. Finally, an \emph{Entropy-Aware Unknown Mixing module} uses both objectness and classification uncertainty to modulate the final class scores in a balanced manner.

\subsection{Incremental LoRA Adapters}
\label{sec:incre_lora}

In EWOD, evolving label space and shifting domain distributions jointly exacerbate severe catastrophic forgetting across tasks. Moreover, since there is no reliance on exemplar replay unlike other methods~\cite{OW-DETR, CAT, PROB, OWOBJ}, the model cannot revisit past data to re-balance its gradients or correct drift from earlier decision boundaries. To tackle these challenges, we propose a \textit{Data-Aware Incremental LoRA Merging} strategy. Specifically, we decouple stability from plasticity by adapting each targeted linear layer in the transformer encoder and decoder using Low-Rank Adaptation (LoRA)~\cite{LoRA}. For each targeted layer at task \(t\) with frozen base weight $\mathbf{W}_0$, we maintain two low-rank adapters:
\begin{itemize}
    \item \textbf{Aggregate LoRA Adapter} \(\big(\Delta \mathbf{W}^{t-1}_{\text{agg}}\big)\): a non-trainable buffer that accumulates knowledge from all previous tasks \(\{\mathcal{T}_1,\ldots,\mathcal{T}_{t-1}\}\), reused in subsequent tasks.
    \item \textbf{Task-Specific LoRA Adapter} \(\big(\Delta \mathbf{W}^{t}_{\text{task}}\big)\): Trainable parameters for the current task \(\mathcal{T}_t\) to capture task-specific shifts in classes/domains, resets at each task transition.
\end{itemize}

\noindent Each adapter is parameterised by a compact memory-efficient product of two low-rank matrices:
\begin{equation}
\Delta \mathbf{W}^{t}_{\text{task}} = \mathbf{B}^{t}_{\text{task}} \mathbf{A}^{t}_{\text{task}}, \qquad
\Delta \mathbf{W}^{t-1}_{\text{agg}} = \mathbf{B}^{t-1}_{\text{agg}} \mathbf{A}^{t-1}_{\text{agg}},
\end{equation}
\noindent During training at task $t$, only the task-specific LoRA matrices $\mathbf{A}^{t}_{\text{task}}$ and $\mathbf{B}^{t}_{\text{task}}$ (plus the detection heads and open-world specific parameters) receive gradients; the base weights and aggregate LoRA adapters are kept frozen.
This design ensures plasticity required for updates specific to $\mathcal{T}_t$ is confined to a small, low-capacity subspace, while the shared representation used across all tasks remains stable.

\noindent A distinctive challenge in EWOD is the severe data imbalance across tasks (details in \texttt{Appendix B}). In such scenarios, the choice of how the Task-Specific and Aggregate LoRA adapters are merged is a key factor that governs the stability–plasticity trade-off: an overly aggressive merge can quickly overwrite useful representations from past tasks, while an overly conservative one can prevent the model from adapting to genuinely new domains. 

\noindent To balance these effects, in EW-DETR, we adopt a deliberately simple, \textbf{data-aware merging coefficient \((\beta_t)\)}, bounded by \(\beta_{\min}\) and \(\beta_{\max}\), that is computed using the ratio between samples seen in current task \((N_t)\) and the cumulative number of samples seen so far previously \((N_{1:t-1})\) (see equation~\ref{eq:beta_t}). As a result, tasks with relatively fewer samples are assigned a larger $\beta_t$, so that under-represented domains exert a stronger influence on the Aggregate LoRA adapter, unlike \cite{CL-LoRA, SD-LoRA}. This keeps the merging rule interpretable while still addressing data imbalance across tasks. 

\begin{equation}
\beta_t =
\begin{cases}
1, & t = 1,\\
\beta_{\max} - (\beta_{\max} - \beta_{\min}) \dfrac{N_t}{N_{1:t-1}}, & t \ge 2.
\end{cases}
\label{eq:beta_t}
\end{equation}

\noindent For task \(t\), the dense merged update is then computed as:
\begin{equation}
\Delta \mathbf{W}^{t}_{\text{merged}} = (1-\beta_t)\,\Delta \mathbf{W}^{t-1}_{\text{agg}} + \beta_t\,\Delta \mathbf{W}^{t}_{\text{task}}.
\end{equation}

\noindent To keep the Aggregate LoRA adapter low-rank, we project this matrix back to rank $r$ using a truncated Singular Value Decomposition (SVD): \(\Delta \mathbf{W}^{t}_{\text{merged}} = \mathbf{U}\,\mathbf{\Sigma}\,\mathbf{V}^\top,\) and retain the top-\(r\) singular components, so that \(\Delta \mathbf{W}^{t}_{\text{merged}} \approx \mathbf{U}_r\,\mathbf{\Sigma}_r\,\mathbf{V}_r^\top\). We store this low-rank approximation and use it to compute Aggregate LoRA adapter for \(\mathcal{T}_t\):
\begin{equation}
    \mathbf{B}^{t}_{\text{agg}} = \mathbf{U}_r\,\mathbf{\Sigma}_r,\qquad
\mathbf{A}^{t}_{\text{agg}} = \mathbf{V}_r^\top,
\end{equation}
\begin{equation}
    \Delta \mathbf{W}_{\text{agg}}^{t} = \mathbf{B}_{\text{agg}}^{t}\mathbf{A}_{\text{agg}}^{t}
\end{equation}
While the Task-Specific LoRA adapter is reset to zero \(\big(\Delta \mathbf{W}^{t}_{\text{task}} \to 0\big)\) in preparation for the next task.  

\subsection{Query-Norm Objectness Adapter}
\label{sec:query_norm_objectness}
In EWOD, there is no explicit supervision for \emph{unknown} objects, so standard classification heads tend to absorb them into the background or misclassify them as known. We therefore introduce the \textbf{Query-Norm Objectness Adapter (QNorm-Obj)}, which decouples the semantics and magnitude of decoder queries: the feature \emph{direction} encodes class semantics, while the feature \emph{norm} acts as a soft, class-agnostic objectness cue. Unlike prior open-world detectors that introduce dedicated objectness or unknown-supervision losses~\cite{ORE, OW-DETR, PROB, ORTH, ZiRA}, QNorm-Obj does not introduce any additional supervision or auxiliary loss.

\noindent In DETR, decoder features \((\mathbf{H})\) can be interpreted as \emph{class-agnostic object slots}~\cite{OV-DETR, labonte2023scaling}: a fixed set of queries attends to image regions, and the classification head maps them to task-specific labels. We exploit these class-agnostic queries to estimate objectness and expose \textit{``unknowns''} in EWOD.

Let \(\mathbf{h}_i\) denote the decoder feature for query \(i\) at the last decoder layer. We first apply LayerNorm followed by $\ell_2$ normalisation to obtain a direction-only vector:
\begin{equation}
  \mathbf{h}_\text{norm} = \frac{\mathrm{LN}(\mathbf{h}_i)}{\|\mathrm{LN}(\mathbf{h}_i)\|_2}.
\end{equation}
The normalised vector lies on the unit sphere and is thus insensitive to changes in feature magnitude that often come from domain-specific covariate shift in EWOD. We then define the classification feature as a convex combination of the original and normalised vectors using a learnable mixing coefficient \(\alpha_{\text{mix}}\):
\begin{equation}
\mathbf{h}_{\text{cls}} = (1-\alpha_{\text{mix}})\mathbf{h}_i + \alpha_{\text{mix}}\mathbf{h}_{\text{norm}},
\label{eq:qnorm_mix}
\end{equation}
and obtain classification logits by:
\begin{equation}
\mathbf{z}_{\text{cls}} = \mathbf{W}_{\text{cls}} \mathbf{h}_{\text{cls}} + \mathbf{b}_{\text{cls}}.
\end{equation}

In DETR-based detectors, queries matched to real objects empirically develop larger norms than unmatched background queries (see \texttt{Appendix C}). QNorm-Obj exploits this by passing the scalar norm \(\|\mathbf{h}_i\|_2\) through an objectness head \(f_{\text{obj}}\) followed by temperature scaling (\(\tau\): temperature, \(\epsilon\): stability constant):
\begin{equation}
z^{\text{obj}}_i = \frac{f_{\text{obj}}\big(\|\mathbf{h}_i\|_2\big)}{\tau + \epsilon}.
\end{equation}
Since no auxiliary loss is introduced, both the normalised classification features and the objectness MLP are trained implicitly through the standard detection loss. Queries matched to ground-truth boxes receive positive gradients on their foreground class and box, encouraging their norms and directions to jointly encode objectness and semantics, while unmatched queries are pushed towards the background. As a result, QNorm-Obj remains compatible with DETR-based detectors, adds negligible overhead, and yields features that are substantially more informative for downstream unknown detection.

\subsection{Entropy-Aware Unknown Mixing}
\label{sec:unknown_mix}

Calibrating the unknown class in EWOD is challenging because unknown instances are never directly labelled, and the detector is trained only on the current known classes. As a result, unknown evidence tends to be either under-confident or spuriously absorbed into nearby known classes by the softmax, especially under domain shift. To address this, we introduce \textbf{Entropy-Aware Unknown Mixing (EUMix)}, which turns classification uncertainty and objectness into a calibrated unknown score.

Given the Query-Norm Objectness Adapter (Sec.~\ref{sec:query_norm_objectness}), each query produces (i) classification logits over the $|\mathcal{K}^t|$ known classes plus a dedicated unknown logit, and (ii) an auxiliary objectness logit. We interpret high objectness as evidence that there is some object at the query location, regardless of class, and the entropy of the known predictions as a measure of how well any known category explains that object. EUMix combines these two signals into: (1) an objectness-driven unknown probability that is high when the detector believes there is an object but all known classes are uncertain, and (2) a classifier-driven unknown probability derived from the learned unknown logit.

These two estimates are blended through a single learnable mixing weight $\alpha \in (0,1)$,
\begin{equation}
p^{\mathrm{unk}}_{\mathrm{final},i}
= \alpha\, p^{\mathrm{unk}}_{\mathrm{cls},i}
+ (1-\alpha)\, p^{\mathrm{unk}}_{\mathrm{obj},i},
\end{equation}
and converted back to a logit that replaces the original unknown logit. In parallel, we apply a soft suppression to the known-class logits proportional to the objectness-driven unknown score, so that high-uncertainty, high-objectness queries are not forced into a known label purely by the softmax normalisation.

All parameters of this module are trained jointly with the detector using the standard detector loss, with no explicit unknown supervision. Unlike prior open-world methods that rely on external energy-based classifiers~\cite{ORE}, proposal memories or OOD proposal banks~\cite{OW-DETR,allabadi2023generalized}, or multi-stage pseudo-labelling pipelines~\cite{ORE,OW-DETR,mullappilly2024semi}, our design operates as a single, lightweight logit-calibration layer on top of the base detector. \texttt{Appendix A} details the mathematical formulation behind EUMix.

\section{Experiments}
\label{sec:experiments}

\subsection{Datasets}
\label{sec:datasets}
Since EWOD requires a series of domain-incremental datasets that also support class evolution across tasks, we follow \cite{DuET} and formulate the EWOD experiments using two dataset series. The Pascal Series comprises four visually distinct domains: Pascal VOC \cite{PASCAL_VOC}, Clipart, Watercolor, and Comic \cite{PASCAL_Series}. Among these, Pascal VOC and Clipart share 20 object classes, while Watercolour and Comic each contain six classes that form subsets of the former. For the Diverse Weather Series, we consider five weather conditions featuring seven common object categories: Daytime Sunny, Night Sunny, Night Rainy, Daytime Foggy, and Dusk Rainy. These are sourced from BDD-100k \cite{BDD100k}, FoggyCityscapes \cite{FoggyCityscapes}, and Adverse-Weather \cite{AdverseWeather}. Detailed dataset statistics are provided in \texttt{Appendix E}.

\subsection{Evaluation Protocol and Metrics}
\label{sec:eval_protocol_main}
\paragraph{Evaluation Protocol.} EWOD requires both: exemplar-free progression over evolving classes \& domains along with explicit unknown modelling. Hence, we build the EWOD protocol by re-formulating the dual-incremental schedule followed in \cite{DuET} for Pascal Series and Diverse Weather benchmarks, while integrating open-world constraints as formulated in OWOD methods \cite{ORE,OW-DETR,OWOBJ}. 

Table S1 illustrates our protocol on the Diverse Weather benchmark. Following OWOD~\cite{ORE,OW-DETR}, during training on task \(\mathcal{T}_t\), only the current task's known classes receive supervision, while instances of previously learned classes and truly novel objects remain unlabeled. During evaluation at \(\mathcal{T}_t\), the detector is supposed to detect all known classes learned so far \(\{\mathcal{K}_1 \cup \ldots \cup \mathcal{K}_t\}\) across all encountered domains \(\{\mathcal{D}_1, \ldots, \mathcal{D}_t\}\) and any unseen objects as \textit{``unknown''} class. Moreover, to create a stationary, non-leaking unknown prior pool, we intentionally withhold certain classes (e.g., \textcolor{red}{\sout{truck}} in Table S1) across all tasks to serve as consistent unknown objects throughout the evaluation, unlike standard OWOD benchmarks: M-OWODB~\cite{ORE} and S-OWODB~\cite{OW-DETR} where there are no unknowns in last task \(\mathcal{T}_T\). 
\paragraph{Evaluation Metrics.} While existing metrics like $\mathcal{F}_{\text{map}}$~\cite{chen2019new} and RAI~\cite{DuET} measure forgetting and generalisation, they do not account for open-set behaviour. Conversely, OWOD metrics~\cite{ORE,OW-DETR} focus on unknown detection but ignore domain shifts. However, EWOD couples these three failure modes: catastrophic forgetting of past classes, collapse under open-set exposure, and lack of domain generalisation ability to identify known classes across evolving domains, where these isolated evaluations might fall short. Hence, to capture these coupled failure modes and facilitate a holistically simpler comparison with a single number, we introduce \textbf{FOGS (Forgetting–Openness–Generalisation Score)}, a comprehensive score that is calculated as the mean of three calibrated sub-scores, each quantifying a distinct EWOD dimension:

\noindent\textbf{Forgetting Sub-Score (FSS)} quantifies catastrophic forgetting by measuring how well the detector retains performance on previously learned classes.  
\begin{equation}
    \text{FSS} = \frac{1}{T-1} \sum_{t=1}^{T-1} \frac{\text{mAP}_{\text{prev}}^{\mathcal{T}_T}}{\text{Avg}(\text{mAP}_{\text{curr}})_{[1:t]}},
    \label{eq:FSS}
\end{equation}
where \(\text{mAP}_{\text{prev}}^{\mathcal{T}_T}\) is the final-task performance on classes learned in task \(t\), and \(\text{Avg}(\text{mAP}_{\text{curr}})_{[1:t]}\) is the average initial performance when those classes were first introduced across tasks $1$ to $t$. Higher FSS indicates better retention of past knowledge.

\begin{table*}[t!]
    \centering
    \caption{EWOD on Pascal Series: VOC [1:10] $\to$ Clipart [11:18]. Along with proposed metrics, per-task OWOD metrics: U-Recall and mAP@0.5 are also reported for completeness. Best results per column in \textbf{bold}, second-best \underline{\textit{underlined}}.}
    \label{tab:E1_extended}
    \vspace{-4mm}
    \renewcommand{\arraystretch}{0.6}
    \resizebox{0.9\textwidth}{!}{%
        \begin{tabular}{l|c||cc|cccc|cccc}
            \toprule
            \multirow{4}{*}{\textbf{Method}} &
            \multirow{4}{*}{\makecell{\textbf{Trainable} \\ \textbf{Params} \\ \textbf{(M)}}} &
            \multicolumn{2}{c|}{\textbf{\(\mathcal{T}_1\): VOC [1:10]}} &
            \multicolumn{4}{c|}{\textbf{\(\mathcal{T}_2\): Clipart [11:18]}} &
            \multicolumn{4}{c}{\textbf{Metrics}} \\
            
            & & 
            \cellcolor{verylightyellow} & 
            \cellcolor{verylightblue}\textbf{mAP ($\uparrow$)} & 
            \cellcolor{verylightyellow} & 
            \multicolumn{3}{c|}{\cellcolor{verylightblue}\textbf{mAP ($\uparrow$)}} & 
            \cellcolor{verylightorange} & 
            \cellcolor{verylightorange} & 
            \cellcolor{verylightorange} & 
            \cellcolor{verylightorange} \\
            
            & & 
            \cellcolor{verylightyellow}\multirow{-3}{*}{\makecell{\textbf{U-Recall} \\ \textbf{($\uparrow$)}}} & 
            \makecell{\textbf{Curr.} \\ \textbf{Known}} & 
            \cellcolor{verylightyellow}\multirow{-3}{*}{\makecell{\textbf{U-Recall} \\ \textbf{($\uparrow$)}}} & 
            \makecell{\textbf{Prev.} \\ \textbf{Known}} & 
            \makecell{\textbf{Curr.} \\ \textbf{Known}} & 
            \textbf{Both} & 
            \cellcolor{verylightorange}\multirow{-3}{*}{\makecell{\textbf{FSS} \\ \textbf{($\uparrow$)}}} & 
            \cellcolor{verylightorange}\multirow{-3}{*}{\makecell{\textbf{OSS} \\ \textbf{($\uparrow$)}}} & 
            \cellcolor{verylightorange}\multirow{-3}{*}{\makecell{\textbf{GSS} \\ \textbf{($\uparrow$)}}} & 
            \cellcolor{verylightorange}\multirow{-3}{*}{\makecell{\textbf{FOGS} \\ \textbf{($\uparrow$)}}} \\ \hline \hline
            
            ORE-EBUI$_{\textcolor{MidnightBlue}{\text{ CVPR'}21}}$~\cite{ORE} & 32.96 & 9.84 & 23.7 & 6.97 & 0 & \underline{\textit{11.37}} & 5.05 & 0 & 55.48 & \underline{\textit{11.37}} & 22.28 \\
            
            OW-DETR$_{\textcolor{MidnightBlue}{\text{ CVPR'}22}}$~\cite{OW-DETR} & 24.22 & 16.25 & 31.51 & 8.07 & 3.6 & 7.96 & 5.54 & 11.42 & 40.47 & 7.96 & 19.95 \\
            
            PROB$_{\textcolor{MidnightBlue}{\text{ CVPR'}23}}$~\cite{PROB} & 23.99 & 52.73 & 12.68 & \underline{\textit{46.27}} & 0 & 0.27 & 0.12 & 0 & \underline{\textit{67.58}} & 0.27 & 22.62 \\
            
            CAT$_{\textcolor{MidnightBlue}{\text{ CVPR'}23}}$~\cite{CAT} & 24.25 & 20.23 & 30.18 & 8.1 & 8.91 & 8.05 & 8.53 & 29.52 & 41.29 & 8.05 & 26.29 \\
            
            ORTH$_{\textcolor{MidnightBlue}{\text{ CVPR'}24}}$~\cite{ORTH} & 105.9 & \underline{\textit{63.92}} & \underline{\textit{67.28}} & 41.61 & 3.92 & \textbf{32.44} & 16.59 & 5.83 & 51.06 & \textbf{32.44} & 29.78 \\
            
            DuET$_{\textcolor{MidnightBlue}{\text{ ICCV'}25}}$~\cite{DuET} & 24.22 & 0 & 34.35 & 0 & 14.1 & 1.46 & 8.47 & 41.05 & 35.49 & 1.46 & 26 \\
            
            OWOBJ$_{\textcolor{MidnightBlue}{\text{ CVPR'}25}}$~\cite{OWOBJ} & 23.99 & 45.93 & 17.47 & 35.72 & 0 & 0.51 & 0.22 & 0 & 60.73 & 0.51 & 20.41 \\ 
            
            \rowcolor{lightgrayblue} \textbf{EW-DETR}$_{\text{ D-DETR}}$ & \textbf{0.46} & 44.98 & 61.64 & 41.45 & \underline{\textit{39.98}} & 7.92 & \underline{\textit{25.73}} & \underline{\textit{64.86}} & 61.67 & 7.92 & \underline{\textit{44.82}} \\
            
            \rowcolor{lightgrayblue} \textbf{EW-DETR}$_{\text{ RF-DETR}}$ & \underline{\textit{1.8}} & \textbf{77.35} & \textbf{76.05} & \textbf{78.23} & \textbf{73.15} & 8.42 & \textbf{45.08} & \textbf{96.19} & \textbf{78.62} & 8.42 & \textbf{61.08} \\ 
            \bottomrule
        \end{tabular}%
    }
    \vspace{-1mm}
\end{table*}
\begin{table*}[t!]
    \centering
    \caption{EWOD results on Pascal Series (left: two-phase) and Diverse Weather (right: multi-phase). Best in \textbf{bold}, second-best \underline{\textit{underlined}}.}
    \label{tab:E2_E3_E9_E10_condensed}
    \vspace{-4mm}
    \resizebox{0.9\textwidth}{!}{%
        \begin{tabular}{l||cccc|cccc|cccc|cccc}
            \toprule
            \multirow{3}{*}{\textbf{Method}} &
            \multicolumn{4}{c|}{\textbf{VOC [1:10] \(\to\) Watercolor [11:14]}} &
            \multicolumn{4}{c|}{\textbf{VOC [1:10] \(\to\) Comic [11:14]}} &
            \multicolumn{4}{c|}{\makecell{\textbf{Daytime Sunny [1:2] \(\to\) Night} \\ \textbf{Sunny [3:4] \(\to\) Night Rainy [5:6]}}} &
            \multicolumn{4}{c}{\makecell{\textbf{Daytime Sunny [1:2] \(\to\) Daytime} \\ \textbf{Foggy [3:4] \(\to\) Dusk Rainy [5:6]}}} \\
            & 
            \cellcolor{verylightorange}\textbf{FSS} & 
            \cellcolor{verylightorange}\textbf{OSS} & 
            \cellcolor{verylightorange}\textbf{GSS} & 
            \cellcolor{verylightorange}\textbf{FOGS} & 
            \cellcolor{verylightorange}\textbf{FSS} & 
            \cellcolor{verylightorange}\textbf{OSS} & 
            \cellcolor{verylightorange}\textbf{GSS} & 
            \cellcolor{verylightorange}\textbf{FOGS} & 
            \cellcolor{verylightorange}\textbf{FSS} & 
            \cellcolor{verylightorange}\textbf{OSS} & 
            \cellcolor{verylightorange}\textbf{GSS} & 
            \cellcolor{verylightorange}\textbf{FOGS} & 
            \cellcolor{verylightorange}\textbf{FSS} & 
            \cellcolor{verylightorange}\textbf{OSS} & 
            \cellcolor{verylightorange}\textbf{GSS} & 
            \cellcolor{verylightorange}\textbf{FOGS} \\ \hline \hline
            
            ORE-EBUI~\cite{ORE} & 0 & 50.7 & 21.46 & 24.05 & 0 & 52.34 & 18.25 & 23.53 & 0 & 61.51 & 18.96 & 26.82 & 0 & 61.14 & \underline{\textit{24.73}} & 28.62 \\
            
            OW-DETR~\cite{OW-DETR} & 11.9 & 41.57 & 10.16 & 21.21 & 1.43 & 41.83 & 8.75 & 17.34 & 0.76 & 61.95 & 21 & 27.9 & 2.31 & 63.02 & 22.23 & 29.19 \\
            
            PROB~\cite{PROB} & 0 & \textbf{63.98} & 3.45 & 22.48 & 0 & \textbf{63.75} & 4.06 & 22.6 & 1.24 & \underline{\textit{66.25}} & 14.8 & 27.43 & 0 & \textbf{66.9} & 12.35 & 26.42 \\
            
            CAT~\cite{CAT} & 1.51 & 44.03 & 12.57 & 23.7 & 7.82 & 43 & 10.31 & 20.38 & 4.01 & 62.92 & 23.66 & 30.2 & 4.85 & 63.87 & 20.5 & 29.74 \\
            
            ORTH~\cite{ORTH} & 12.32 & 46.89 & \textbf{45.01} & 34.74 & 5.28 & 51.56 & \textbf{38.49} & 31.78 & 0.08 & 62.68 & \textbf{40.38} & \underline{\textit{34.38}} & 0.31 & 61.81 & \textbf{48.56} & \underline{\textit{36.89}} \\
            
            DuET~\cite{DuET} & 46.7 & 33.01 & 4.67 & 28.13 & 56.22 & 35 & 4.99 & 32.07 & 20.66 & 56.47 & 2.68 & 26.6 & \underline{\textit{29.01}} & 55.42 & 3.97 & 29.47 \\
            
            OWOBJ~\cite{OWOBJ} & 0 & 56.6 & 18.37 & 24.99 & 0 & \underline{\textit{63.11}} & 4.97 & 22.69 & 0 & 64.22 & 12.09 & 25.44 & 0 & 55.59 & 9.02 & 21.54 \\ 
            
            \rowcolor{lightgrayblue} \textbf{EW-DETR}$_{\text{D-DETR}}$ & \underline{\textit{79.61}} & 54.46 & 19.99 & \underline{\textit{51.35}} & \underline{\textit{83.18}} & 54.38 & 4.19 & \underline{\textit{47.25}} & \underline{\textit{26.84}} & 56.09 & \underline{\textit{19.86}} & 34.26 & 20.55 & 59.58 & 20.86 & 33.66 \\
            
            \rowcolor{lightgrayblue} \textbf{EW-DETR}$_{\text{RF-DETR}}$ & \textbf{98.96} & \underline{\textit{58.17}} & \underline{\textit{40.5}} & \textbf{65.88} & \textbf{98.51} & 56.68 & \underline{\textit{32.91}} & \textbf{62.7} & \textbf{73.63} & \textbf{73.43} & 18.68 & \textbf{55.25} & \textbf{82.80} & \underline{\textit{65.75}} & 18.13 & \textbf{55.56} \\ 
            \bottomrule
        \end{tabular}
    }
    \vspace{-1mm}
\end{table*}
\begin{table*}[t!]
    \centering
    \caption{EWOD on Diverse Weather Series: Daytime Sunny $\to$ Night/Foggy/Rainy transitions. Best in \textbf{bold}, second-best \underline{\textit{underlined}}.}
    \label{tab:E4-E7_condensed}
    \vspace{-4mm}
    \resizebox{0.9\textwidth}{!}{
        \begin{tabular}{l||cccc|cccc|cccc|cccc}
            \toprule
            \multirow{2}{*}{\textbf{Method}} & 
            \multicolumn{4}{c|}{\makecell{\textbf{Daytime Sunny [1:3]} \\ \textbf{$\to$ Night Sunny [4:6]}}} & 
            \multicolumn{4}{c|}{\makecell{\textbf{Daytime Sunny [1:3]} \\ \textbf{$\to$ Night Rainy [4:6]}}} & 
            \multicolumn{4}{c|}{\makecell{\textbf{Daytime Sunny [1:3]} \\ \textbf{$\to$ Daytime Foggy[4:6]}}} & 
            \multicolumn{4}{c}{\makecell{\textbf{Daytime Sunny [1:3]} \\ \textbf{$\to$ Dusk Rainy [4:6]}}} \\
            
            & 
            \cellcolor{verylightorange}\textbf{FSS} & 
            \cellcolor{verylightorange}\textbf{OSS} & 
            \cellcolor{verylightorange}\textbf{GSS} & 
            \cellcolor{verylightorange}\textbf{FOGS} & 
            \cellcolor{verylightorange}\textbf{FSS} & 
            \cellcolor{verylightorange}\textbf{OSS} & 
            \cellcolor{verylightorange}\textbf{GSS} & 
            \cellcolor{verylightorange}\textbf{FOGS} & 
            \cellcolor{verylightorange}\textbf{FSS} & 
            \cellcolor{verylightorange}\textbf{OSS} & 
            \cellcolor{verylightorange}\textbf{GSS} & 
            \cellcolor{verylightorange}\textbf{FOGS} & 
            \cellcolor{verylightorange}\textbf{FSS} & 
            \cellcolor{verylightorange}\textbf{OSS} & 
            \cellcolor{verylightorange}\textbf{GSS} & 
            \cellcolor{verylightorange}\textbf{FOGS} \\ \hline \hline
            
            ORE-EBUI~\cite{ORE} & 0 & 61.06 & 20.41 & 27.16 & 0 & 61.44 & \textit{\underline{9.18}} & 23.54 & 0 & 60.35 & 24.36 & 28.24 & 0 & 61.51 & \textit{\underline{16.45}} & 25.99 \\
            
            OW-DETR~\cite{OW-DETR} & 0 & 61.06 & 23 & 28.02 & 6.54 & 61.79 & 6.84 & 25.06 & 0 & 61.49 & 23.39 & 28.29 & 0.71 & 60.39 & 12.47 & 24.52 \\
            
            PROB~\cite{PROB} & 0 & \textit{\underline{69.01}} & 8.25 & 25.75 & 0.96 & \textit{\underline{69.35}} & 3.98 & 24.76 & 0 & \textbf{69.49} & 10.57 & 26.69 & 0 & \textit{\underline{69.15}} & 11.34 & 26.83 \\
            
            CAT~\cite{CAT} & 0 & 60.77 & \textit{\underline{24.43}} & 28.4 & 7.71 & 61.47 & 7.9 & 25.69 & 0 & 60.72 & \textit{\underline{29.02}} & 29.91 & 1.1 & 60.85 & 12.41 & 24.56 \\
            
            ORTH~\cite{ORTH} & 0.31 & 59.33 & \textbf{40.56} & \textit{\underline{33.4}} & 2.61 & 64.21 & \textbf{22.81} & 29.88 & 0.54 & 60.94 & \textbf{52.89} & \textit{\underline{38.12}} & 0.49 & 67.88 & \textbf{36.31} & \textit{\underline{34.89}} \\
            
            DuET~\cite{DuET} & 3.07 & 55.07 & 0.75 & 19.63 & 16.45 & 53.78 & 0.49 & 23.57 & \textit{\underline{20.41}} & 50.36 & 8.6 & 26.46 & 5.84 & 52.79 & 0.87 & 19.83 \\
            
            OWOBJ~\cite{OWOBJ} & 0 & 67.4 & 9.17 & 25.52 & 0 & 62.27 & 1.99 & 21.42 & 0 & \textit{\underline{64.57}} & 16.61 & 27.06 & 0 & 64.59 & 6.72 & 23.77 \\
            
            \rowcolor{lightgrayblue} \textbf{EW-DETR}$_{\text{D-DETR}}$ & \textit{\underline{22.85}} & 56.03 & 10.35 & 29.74 & \textit{\underline{31.74}} & 55.44 & 6.81 & \textit{\underline{31.33}} & 14.36 & 57.39 & 20.01 & 30.59 & \textit{\underline{14.5}} & 57.08 & 14.09 & 28.56 \\
            
            \rowcolor{lightgrayblue} \textbf{EW-DETR}$_{\text{RF-DETR}}$ & \textbf{54.47} & \textbf{71.55} & 15.78 & \textbf{47.27} & \textbf{85.25} & \textbf{69.68} & 3.35 & \textbf{52.76} & \textbf{58.93} & 64.25 & 22.18 & \textbf{48.45} & \textbf{67.12} & \textbf{73.56} & 15.42 & \textbf{52.03} \\ 
            \bottomrule
        \end{tabular}
    }
    \vspace{-4mm}
\end{table*}
\noindent\textbf{Openness Sub-Score (OSS)} captures the open-set behaviour by combining three unknown detection metrics from OWOD works \cite{ORE,OW-DETR,OWOBJ}:
\begin{equation}
\text{OSS} = \frac{1}{T} \sum_{t=1}^{T} \frac{\text{U-Recall}_t + (1-\text{WI}_t) + \frac{1}{1 + \text{A-OSE}_t/\text{GT}_{\text{unk},t}}}{3},
\label{eq:OSS}
\end{equation}
where U-Recall measures the detector's ability to recover unknown objects, \((1-\text{WI})\) rewards resisting precision collapse when unknowns are present (WI: Wilderness Impact~\cite{WI}), and the normalised A-OSE (Absolute Open-Set Error~\cite{A-OSE}) penalizes misclassification of unknowns as known classes with respect to ground-truth unknowns \(\text{GT}_{\text{unk},t}\) at task \(t\). Higher OSS indicates robust unknown detection.

\noindent\textbf{Generalisation Sub-Score (GSS)} evaluates cross-domain adaptability and measures how well newly learned classes transfer across the mixed domains seen so far.
\begin{equation}
\text{GSS} = \frac{1}{T-1} \sum_{t=2}^{T} \text{mAP}_{\text{curr}}^{\mathcal{T}_t},
\label{eq:GSS}
\end{equation}
where $\text{mAP}_{\text{curr}}^{\mathcal{T}_t}$ is the performance on current task classes when evaluated across the union of all domains seen up to task $t$. Higher GSS reflects stronger domain generalisation.

Detailed metric formulations are provided in \texttt{Appendix D}.

\section{Results \& Discussion}
\label{sec:results}
\begin{figure*}[ht]
\centerline{\includegraphics[width=\textwidth]{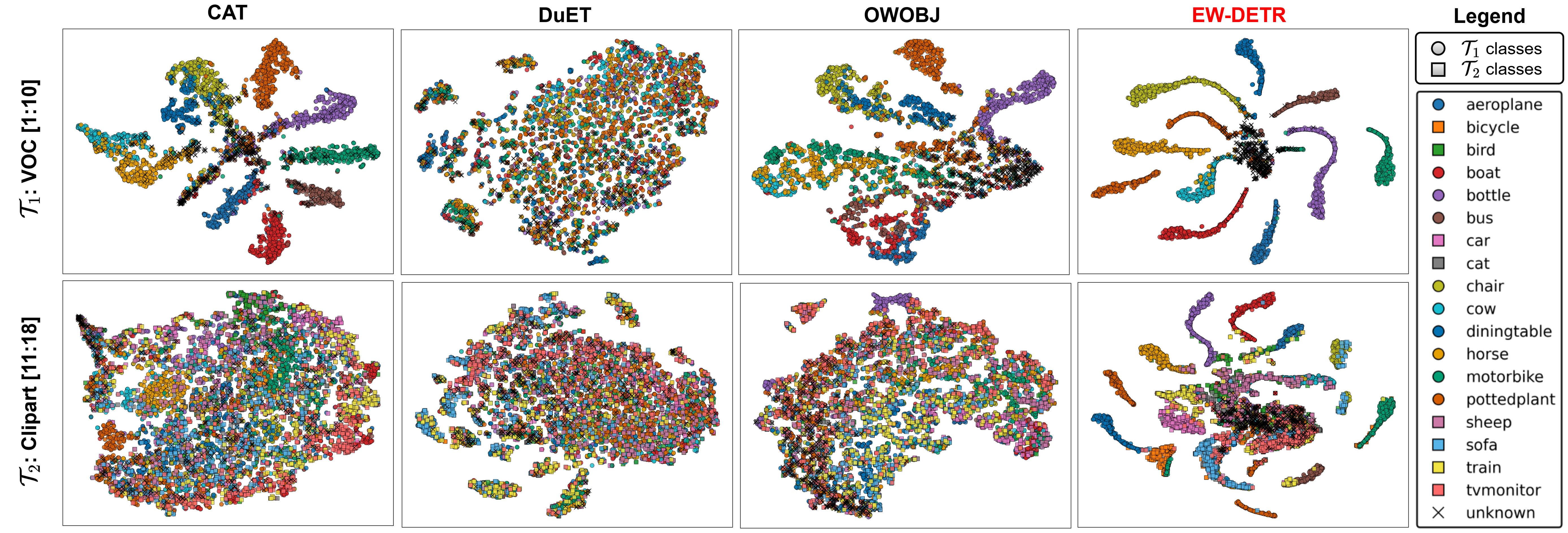}}
\vspace{-0.5em}
\caption{\textbf{t-SNE visualisation of decoder query features comparing EW-DETR with recent methods on VOC [1:10] \(\to\) Clipart [11:18] task.} The top row shows features obtained during inference on VOC [1:10] classes, while the bottom row shows features evaluated on the combined test set comprising VOC [1:18] and Clipart [1:18] classes. Notably, EW-DETR is only able to preserve distinct class clusters even under severe domain shift (VOC \(\to\) Clipart) while also generalising to unseen class-domain pairs: VOC [11:18] and Clipart [1:10], whereas other recent methods exhibit severe feature collapse in \(\mathcal{T}_2\). Best viewed in colour with zoom.}
\vspace{-2mm}
\label{fig:tSNE_results}
\end{figure*}
\begin{table*}[ht]
\centering
\caption{Ablations of proposed modules of the EW-DETR framework on RF-DETR~\cite{RF-DETR} for: VOC [1:10] $\to$ Clipart [11:18] benchmark.}
\label{tab:ablations}
\vspace{-2mm}
\renewcommand{\arraystretch}{1.2}
\resizebox{\textwidth}{!}{
\begin{tabular}{l||cc|cccc|cccc|c}
\toprule

\multirow{4}{*}{\textbf{Configuration}} & \multicolumn{2}{c|}{\textbf{\(\mathcal{T}_1\): VOC [1:10]}} & \multicolumn{4}{c|}{\textbf{\(\mathcal{T}_2\): Clipart [11:18]}} & \multicolumn{4}{c|}{\textbf{Metrics}} & \multirow{4}{*}{\makecell{\textbf{Trainable} \\ \textbf{Params} \\ \textbf{(M)}}} \\ 

& \cellcolor{verylightyellow} & \cellcolor{verylightblue}\textbf{mAP ($\uparrow$)} & \cellcolor{verylightyellow} & \multicolumn{3}{c|}{\cellcolor{verylightblue}\textbf{mAP ($\uparrow$)}} & 
\cellcolor{verylightorange} & \cellcolor{verylightorange} & \cellcolor{verylightorange} & \cellcolor{verylightorange} & \\

& \cellcolor{verylightyellow}\multirow{-2}{*}{\makecell{\textbf{U-Recall} \\ \textbf{($\uparrow$)}}} & \makecell{\textbf{Curr.} \\ \textbf{Known}} & \cellcolor{verylightyellow}\multirow{-2}{*}{\makecell{\textbf{U-Recall} \\ \textbf{($\uparrow$)}}} & \makecell{\textbf{Prev.} \\ \textbf{Known}} & \makecell{\textbf{Curr.} \\ \textbf{Known}} & \textbf{Both} & 
\cellcolor{verylightorange}\multirow{-2}{*}{\makecell{\textbf{FSS} \\ \textbf{($\uparrow$)}}} & 
\cellcolor{verylightorange}\multirow{-2}{*}{\makecell{\textbf{OSS} \\ \textbf{($\uparrow$)}}} & 
\cellcolor{verylightorange}\multirow{-2}{*}{\makecell{\textbf{GSS} \\ \textbf{($\uparrow$)}}} & 
\cellcolor{verylightorange}\multirow{-2}{*}{\makecell{\textbf{FOGS} \\ \textbf{($\uparrow$)}}} & \\ \hline \hline

Baseline & 1.46 & 77.21 & 0.56 & 5.81 & 51.49 & 26.11 & 7.52 & 33.78 & 51.49 & 30.93 & 30.8672 \\ 
+ Train \(\Delta\mathbf{W}^{t}_{\text{task}}\) only (simple PEFT, no agg.) & 1.20 & 77.23 & 0.84 & 20.48 & 52.33 & 34.63 & 26.52 & 33.34 & 52.33 & 37.4 & 1.7997 \\ 
+ Train \(\Delta\mathbf{W}^{t-1}_{\text{agg}}~\&~\Delta\mathbf{W}^{t}_{\text{task}}\) (dual LoRA; no merging/SVD) & 1.77 & 77.07 & 0.11 & 20.59 & 50.49 & 33.87 & 26.72 & 34.26 & 50.49 & 37.16 & 1.7997 \\ 
+ Train Incre. LoRA & 1.08 & 76.29 & 0.29 & 74.85 & 0.07 & 41.61 & 98.11 & 33.53 & 0.07 & 43.9 & 1.7997 \\ 
+ Train QNorm-Obj only & 3.17 & 77.15 & 3.23 & 6.2 & 51.2 & 26.2 & 8.04 & 34.18 & 51.20 & 31.14 & 30.8679 \\
+ Train Incre. LoRA + QNorm-Obj & 4.24 & 76.11 & 2.53 & 74.42 & 5.07 & 42.89 & 97.78 & 42.04 & 5.07 & 48.3 & 1.8004 \\ 
+ Train Incre. LoRA + QNorm-Obj + EUMix & 77.35 & 76.05 & 78.23 & 73.15 & 8.42 & 45.08 & 96.19 & 78.62 & 8.42 & 61.08 & 1.8004 \\ \bottomrule
\end{tabular}
}
\vspace{-4mm}
\end{table*}

\paragraph{Quantitative Results:}
\label{sec:quant_results}
We compare EW-DETR against the recent OWOBJ~\cite{OWOBJ} and DuET~\cite{DuET}, as well as earlier approaches OW-DETR~\cite{OW-DETR}, PROB~\cite{PROB}, CAT~\cite{CAT}, ORTH~\cite{ORTH}, and the pioneering ORE~\cite{ORE}, all re-implemented under our EWOD protocol, as shown in Tables~\ref{tab:E1_extended}, ~\ref{tab:E2_E3_E9_E10_condensed} and ~\ref{tab:E4-E7_condensed}. Standard replay-based OWOD methods (ORE, OW-DETR, CAT, OWOBJ) suffer from severe catastrophic forgetting, with Forgetting scores near zero. PROB~\cite{PROB} attains strong open-set performance (OSS) but fails in retention and generalisation. ORTH~\cite{ORTH} shows strong generalisation yet poor retention, while DuET~\cite{DuET} retains some knowledge but performs weakly under open-world settings. In contrast, EW-DETR delivers balanced performance, achieving the highest average retention (FSS: \textbf{75.69}) with comparatively fewer parameters, the best open-set performance (OSS: \textbf{67.3}), and competitive generalisation (GSS: \textbf{14.02}). Overall, EW-DETR achieves the highest average FOGS score of \textbf{52.33}, making it the most feasible solution for EWOD. Comprehensive results, including additional metrics (WI~\cite{WI} and A-OSE~\cite{A-OSE}) that are also part of OSS, are provided in \texttt{Appendix G4}.

\paragraph{Qualitative Visualizations}
\label{sec:qual_analysis}
To evaluate semantic preservation and separation across domains, we visualise t-SNE embeddings of decoder query features in Figure~\ref{fig:tSNE_results}. All methods form distinct clusters in \(\mathcal{T}_1\), where evaluation is confined to VOC [1:10] classes. However, the true challenge lies in \(\mathcal{T}_2\), where evaluation involves all seen classes across both domains: VOC [1:18] and Clipart [1:18], including class–domain pairs unseen during training (VOC [11:18] \& Clipart [1:10]), which effectively capture the domain generalisation ability of the detector. Only EW-DETR preserves old class clusters while distinctly separating new ones under severe domain shifts (as also evident in Table~\ref{tab:E1_extended}), demonstrating the effectiveness of the proposed modules. Further visualisations are provided in \texttt{Appendix G5}.

\section{Ablations}
\label{sec:ablations}
This section covers the component-wise ablation studies of each proposed module in EW-DETR, while additional ablations, including hyperparameter sensitivity analysis, stability under random class-domain sequences and computational complexity analysis, are provided in \texttt{Appendix G}.

\vspace{-1em}
\paragraph{Incremental LoRA Adapters.} As shown in Table~\ref{tab:ablations}, Incremental LoRA Adapters significantly mitigate catastrophic forgetting, boosting FSS from 7.52 to 98.11. This is because of the substantial recovery of previous known classes in Task $\mathcal{T}_2$ (mAP 5.81 $\rightarrow$ 74.85). Moreover, as discussed in Section~\ref{sec:incre_lora}, base model and transformer encoder-decoder weights are frozen, along with the Aggregate LoRA adapters; this effectively reduces trainable parameters by \textbf{94.2\%}. However, this stability severely costs plasticity, as the $\mathcal{T}_2$ 'Curr. Known' mAP drops from 51.49 to 0.07.

\vspace{-1em}
\paragraph{Query-Norm Objectness Adapter.} Since QNorm-Obj decouples objectness-aware features, incorporating it partially mitigates the open-set deficiency, as indicated by minor increases in U-Recall across both tasks. More importantly, this module preserves strong forgetting resistance (FSS remains high at 97.78) while enhancing generalisation (GSS), which results from the rise in the current known mAP in \(\mathcal{T}_2\).

\vspace{-1em}
\paragraph{Entropy-Aware Unknown Mixing.} Table~\ref{sec:ablations} depicts EUMix works synergistically with the previous two modules and not only improves unknown detection but also enhances the model's ability to generalise current task knowledge, while slightly reducing FSS.  

\section{Conclusion}
\label{sec:conclusion}
In this work, we formalised Evolving World Object Detection to solve real-world coupled challenges of open-set recognition and domain-adaptive incremental learning under exemplar-free constraints. To address these challenges, we proposed EW-DETR, which integrates Data-Aware Incremental LoRA Merging for domain plasticity, Query-Norm Objectness Adapter for robust unknown detection, and Entropy-Aware Unknown Mixing for calibrated open-set recognition. Extensive experiments on two challenging benchmarks demonstrated that EW-DETR significantly outperforms existing methods across all EWOD dimensions. Qualitative analyses further confirmed its ability to maintain discriminative, domain-invariant representations without replay. We believe EW-DETR opens a promising direction for real-world object detection in dynamic environments, and we hope our work inspires further research at the intersection of open-world recognition and domain-adaptive continual learning.

{
    \clearpage
    \small
    \bibliographystyle{ieeenat_fullname}
    \bibliography{main}

@String(ICCV= {Int. Conf. Comput. Vis.})

@String(ICLR = {Int. Conf. Learn. Represent.})

@String(AAAI = {AAAI})

@String(ICCV  = {ICCV})

@String(ICLR  = {ICLR})

@InProceedings{DuET,
    author    = {Monga, Munish and Chudasama, Vishal and Wasnik, Pankaj and Banerjee, Biplab},
    title     = {DuET: Dual Incremental Object Detection via Exemplar-Free Task Arithmetic},
    booktitle = {Proceedings of the IEEE/CVF International Conference on Computer Vision (ICCV)},
    month     = {October},
    year      = {2025},
    pages     = {3121-3131}
}

@inproceedings{ORE,
  title={Towards open world object detection},
  author={Joseph, KJ and Khan, Salman and Khan, Fahad Shahbaz and Balasubramanian, Vineeth N},
  booktitle={Proceedings of the IEEE/CVF conference on computer vision and pattern recognition},
  pages={5830--5840},
  year={2021}
}

@inproceedings{PROB,
  title={Prob: Probabilistic objectness for open world object detection},
  author={Zohar, Orr and Wang, Kuan-Chieh and Yeung, Serena},
  booktitle={Proceedings of the IEEE/CVF Conference on Computer Vision and Pattern Recognition},
  pages={11444--11453},
  year={2023}
}

@article{D-DETR,
  title={Deformable detr: Deformable transformers for end-to-end object detection},
  author={Zhu, Xizhou and Su, Weijie and Lu, Lewei and Li, Bin and Wang, Xiaogang and Dai, Jifeng},
  journal={arXiv preprint arXiv:2010.04159},
  year={2020}
}

@misc{RF-DETR,
    title={RF-DETR: Neural Architecture Search for Real-Time Detection Transformers},
    author={Isaac Robinson and Peter Robicheaux and Matvei Popov and Deva Ramanan and Neehar Peri},
    year={2025},
    eprint={2511.09554},
    archivePrefix={arXiv},
    primaryClass={cs.CV},
    url={https://arxiv.org/abs/2511.09554},
}

@inproceedings{DETR,
  title={End-to-end object detection with transformers},
  author={Carion, Nicolas and Massa, Francisco and Synnaeve, Gabriel and Usunier, Nicolas and Kirillov, Alexander and Zagoruyko, Sergey},
  booktitle={European conference on computer vision},
  pages={213--229},
  year={2020},
  organization={Springer}
}

@inproceedings{OWOBJ,
  title={Open-World Objectness Modeling Unifies Novel Object Detection},
  author={Zhang, Shan and Ni, Yao and Du, Jinhao and Xue, Yuan and Torr, Philip and Koniusz, Piotr and van den Hengel, Anton},
  booktitle={Proceedings of the Computer Vision and Pattern Recognition Conference},
  pages={30332--30342},
  year={2025}
}

@inproceedings{RN50,
  title={Deep residual learning for image recognition},
  author={He, Kaiming and Zhang, Xiangyu and Ren, Shaoqing and Sun, Jian},
  booktitle={Proceedings of the IEEE conference on computer vision and pattern recognition},
  pages={770--778},
  year={2016}
}

@inproceedings{OV-DETR,
  title={Open-vocabulary detr with conditional matching},
  author={Zang, Yuhang and Li, Wei and Zhou, Kaiyang and Huang, Chen and Loy, Chen Change},
  booktitle={European conference on computer vision},
  pages={106--122},
  year={2022},
  organization={Springer}
}

@inproceedings{labonte2023scaling,
  title={Scaling novel object detection with weakly supervised detection transformers},
  author={LaBonte, Tyler and Song, Yale and Wang, Xin and Vineet, Vibhav and Joshi, Neel},
  booktitle={Proceedings of the IEEE/CVF winter conference on applications of computer vision},
  pages={85--96},
  year={2023}
}

@article{LoRA,
  title={Lora: Low-rank adaptation of large language models.},
  author={Hu, Edward J and Shen, Yelong and Wallis, Phillip and Allen-Zhu, Zeyuan and Li, Yuanzhi and Wang, Shean and Wang, Lu and Chen, Weizhu and others},
  journal={ICLR},
  volume={1},
  number={2},
  pages={3},
  year={2022}
}

@inproceedings{OW-DETR,
  title={Ow-detr: Open-world detection transformer},
  author={Gupta, Akshita and Narayan, Sanath and Joseph, KJ and Khan, Salman and Khan, Fahad Shahbaz and Shah, Mubarak},
  booktitle={Proceedings of the IEEE/CVF conference on computer vision and pattern recognition},
  pages={9235--9244},
  year={2022}
}

@inproceedings{CAT,
  title={Cat: Localization and identification cascade detection transformer for open-world object detection},
  author={Ma, Shuailei and Wang, Yuefeng and Wei, Ying and Fan, Jiaqi and Li, Thomas H and Liu, Hongli and Lv, Fanbing},
  booktitle={Proceedings of the IEEE/CVF Conference on Computer Vision and Pattern Recognition},
  pages={19681--19690},
  year={2023}
}

@article{allabadi2023generalized,
  title={Generalized Open-World Semi-Supervised Object Detection},
  author={Allabadi, Garvita and Lucic, Ana and Aananth, Siddarth and Yang, Tiffany and Wang, Yu-Xiong and Adve, Vikram},
  journal={arXiv preprint arXiv:2307.15710},
  year={2023}
}

@inproceedings{mullappilly2024semi,
  title={Semi-supervised open-world object detection},
  author={Mullappilly, Sahal Shaji and Gehlot, Abhishek Singh and Anwer, Rao Muhammad and Khan, Fahad Shahbaz and Cholakkal, Hisham},
  booktitle={Proceedings of the AAAI Conference on Artificial Intelligence},
  volume={38},
  number={5},
  pages={4305--4314},
  year={2024}
}

@article{zhong2025replay,
  title={Replay-Based Incremental Object Detection With Local Response Exploration},
  author={Zhong, Jian and Jiao, Yifan and Bao, Bing-Kun},
  journal={IEEE Transactions on Multimedia},
  year={2025},
  publisher={IEEE}
}

@inproceedings{SDDGR,
  title={Sddgr: Stable diffusion-based deep generative replay for class incremental object detection},
  author={Kim, Junsu and Cho, Hoseong and Kim, Jihyeon and Tiruneh, Yihalem Yimolal and Baek, Seungryul},
  booktitle={Proceedings of the IEEE/CVF Conference on Computer Vision and Pattern Recognition},
  pages={28772--28781},
  year={2024}
}

@inproceedings{CL-DETR,
  title={Continual detection transformer for incremental object detection},
  author={Liu, Yaoyao and Schiele, Bernt and Vedaldi, Andrea and Rupprecht, Christian},
  booktitle={Proceedings of the IEEE/CVF Conference on Computer Vision and Pattern Recognition},
  pages={23799--23808},
  year={2023}
}

@inproceedings{Rodeo,
title={RODEO: Replay for Online Object Detection},
author={Acharya, Manoj and Hayes, Tyler L. and Kanan, Christopher},
booktitle={The British Machine Vision Conference},
year={2020}
}

@inproceedings{LDB,
  title={Non-exemplar Domain Incremental Object Detection via Learning Domain Bias},
  author={Song, Xiang and He, Yuhang and Dong, Songlin and Gong, Yihong},
  booktitle={Proceedings of the AAAI Conference on Artificial Intelligence},
  volume={38},
  number={13},
  pages={15056--15065},
  year={2024}
}

@article{DINO,
  title={Dino: Detr with improved denoising anchor boxes for end-to-end object detection},
  author={Zhang, Hao and Li, Feng and Liu, Shilong and Zhang, Lei and Su, Hang and Zhu, Jun and Ni, Lionel M and Shum, Heung-Yeung},
  journal={arXiv preprint arXiv:2203.03605},
  year={2022}
}

@article{DINOv2,
  title={Dinov2: Learning robust visual features without supervision},
  author={Oquab, Maxime and Darcet, Timoth{\'e}e and Moutakanni, Th{\'e}o and Vo, Huy and Szafraniec, Marc and Khalidov, Vasil and Fernandez, Pierre and Haziza, Daniel and Massa, Francisco and El-Nouby, Alaaeldin and others},
  journal={arXiv preprint arXiv:2304.07193},
  year={2023}
}

@article{verma2023privacy,
  title={Privacy-preserving continual learning methods for medical image classification: a comparative analysis},
  author={Verma, Tanvi and Jin, Liyuan and Zhou, Jun and Huang, Jia and Tan, Mingrui and Choong, Benjamin Chen Ming and Tan, Ting Fang and Gao, Fei and Xu, Xinxing and Ting, Daniel S and others},
  journal={Frontiers in Medicine},
  volume={10},
  pages={1227515},
  year={2023},
  publisher={Frontiers Media SA}
}

@article{PASCAL_VOC,
  title={The pascal visual object classes (voc) challenge},
  author={Everingham, Mark and Van Gool, Luc and Williams, Christopher KI and Winn, John and Zisserman, Andrew},
  journal={International journal of computer vision},
  volume={88},
  pages={303--338},
  year={2010},
  publisher={Springer}
}

@inproceedings{PASCAL_Series,
  title={Cross-domain weakly-supervised object detection through progressive domain adaptation},
  author={Inoue, Naoto and Furuta, Ryosuke and Yamasaki, Toshihiko and Aizawa, Kiyoharu},
  booktitle={Proceedings of the IEEE conference on computer vision and pattern recognition},
  pages={5001--5009},
  year={2018}
}

@inproceedings{BDD100k,
  title={Bdd100k: A diverse driving dataset for heterogeneous multitask learning},
  author={Yu, Fisher and Chen, Haofeng and Wang, Xin and Xian, Wenqi and Chen, Yingying and Liu, Fangchen and Madhavan, Vashisht and Darrell, Trevor},
  booktitle={Proceedings of the IEEE/CVF conference on computer vision and pattern recognition},
  pages={2636--2645},
  year={2020}
}

@inproceedings{FoggyCityscapes,
  title={The cityscapes dataset for semantic urban scene understanding},
  author={Cordts, Marius and Omran, Mohamed and Ramos, Sebastian and Rehfeld, Timo and Enzweiler, Markus and Benenson, Rodrigo and Franke, Uwe and Roth, Stefan and Schiele, Bernt},
  booktitle={Proceedings of the IEEE conference on computer vision and pattern recognition},
  pages={3213--3223},
  year={2016}
}

@article{AdverseWeather,
  title={Vehicle detection and tracking in adverse weather using a deep learning framework},
  author={Hassaballah, Mahmoud and Kenk, Mourad A and Muhammad, Khan and Minaee, Shervin},
  journal={IEEE transactions on intelligent transportation systems},
  volume={22},
  number={7},
  pages={4230--4242},
  year={2020},
  publisher={IEEE}
}

@inproceedings{WI,
  title={The overlooked elephant of object detection: Open set},
  author={Dhamija, Akshay and Gunther, Manuel and Ventura, Jonathan and Boult, Terrance},
  booktitle={Proceedings of the IEEE/CVF winter conference on applications of computer vision},
  pages={1021--1030},
  year={2020}
}

@inproceedings{A-OSE,
  title={Dropout sampling for robust object detection in open-set conditions},
  author={Miller, Dimity and Nicholson, Lachlan and Dayoub, Feras and S{\"u}nderhauf, Niko},
  booktitle={2018 IEEE International Conference on Robotics and Automation (ICRA)},
  pages={3243--3249},
  year={2018},
  organization={IEEE}
}

@inproceedings{chen2019new,
  title={A new knowledge distillation for incremental object detection},
  author={Chen, Li and Yu, Chunyan and Chen, Lvcai},
  booktitle={2019 International Joint Conference on Neural Networks (IJCNN)},
  pages={1--7},
  year={2019},
  organization={IEEE}
}

@inproceedings{ORTH,
  title={Exploring orthogonality in open world object detection},
  author={Sun, Zhicheng and Li, Jinghan and Mu, Yadong},
  booktitle={Proceedings of the IEEE/CVF conference on computer vision and pattern recognition},
  pages={17302--17312},
  year={2024}
}

@inproceedings{pang2019libra,
  title={Libra r-cnn: Towards balanced learning for object detection},
  author={Pang, Jiangmiao and Chen, Kai and Shi, Jianping and Feng, Huajun and Ouyang, Wanli and Lin, Dahua},
  booktitle={Proceedings of the IEEE/CVF conference on computer vision and pattern recognition},
  pages={821--830},
  year={2019}
}

@article{Faster_ILOD,
  title={Faster ilod: Incremental learning for object detectors based on faster rcnn},
  author={Peng, Can and Zhao, Kun and Lovell, Brian C},
  journal={Pattern recognition letters},
  volume={140},
  pages={109--115},
  year={2020},
  publisher={Elsevier}
}

@InProceedings{ABR_IOD,
    author    = {Liu, Yuyang and Cong, Yang and Goswami, Dipam and Liu, Xialei and van de Weijer, Joost},
    title     = {Augmented Box Replay: Overcoming Foreground Shift for Incremental Object Detection},
    booktitle = {Proceedings of the IEEE/CVF International Conference on Computer Vision (ICCV)},
    month     = {October},
    year      = {2023},
    pages     = {11367-11377}
}

@inproceedings{cermelli2022modeling,
  title={Modeling missing annotations for incremental learning in object detection},
  author={Cermelli, Fabio and Geraci, Antonino and Fontanel, Dario and Caputo, Barbara},
  booktitle={Proceedings of the IEEE/CVF Conference on Computer Vision and Pattern Recognition},
  pages={3700--3710},
  year={2022}
}

@inproceedings{shmelkov2017incremental,
  title={Incremental learning of object detectors without catastrophic forgetting},
  author={Shmelkov, Konstantin and Schmid, Cordelia and Alahari, Karteek},
  booktitle={Proceedings of the IEEE international conference on computer vision},
  pages={3400--3409},
  year={2017}
}

@article{SID,
  title={SID: incremental learning for anchor-free object detection via selective and inter-related distillation},
  author={Peng, Can and Zhao, Kun and Maksoud, Sam and Li, Meng and Lovell, Brian C},
  journal={Computer vision and image understanding},
  volume={210},
  pages={103229},
  year={2021},
  publisher={Elsevier}
}

@inproceedings{RILOD,
  title={RILOD: Near real-time incremental learning for object detection at the edge},
  author={Li, Dawei and Tasci, Serafettin and Ghosh, Shalini and Zhu, Jingwen and Zhang, Junting and Heck, Larry},
  booktitle={Proceedings of the 4th ACM/IEEE Symposium on Edge Computing},
  pages={113--126},
  year={2019}
}

@inproceedings{ERD,
  title={Overcoming catastrophic forgetting in incremental object detection via elastic response distillation},
  author={Feng, Tao and Wang, Mang and Yuan, Hangjie},
  booktitle={Proceedings of the IEEE/CVF conference on computer vision and pattern recognition},
  pages={9427--9436},
  year={2022}
}

@article{S-prompts,
  title={S-prompts learning with pre-trained transformers: An occam’s razor for domain incremental learning},
  author={Wang, Yabin and Huang, Zhiwu and Hong, Xiaopeng},
  journal={Advances in Neural Information Processing Systems},
  volume={35},
  pages={5682--5695},
  year={2022}
}

@inproceedings{DISC,
  title={An efficient domain-incremental learning approach to drive in all weather conditions},
  author={Mirza, M Jehanzeb and Masana, Marc and Possegger, Horst and Bischof, Horst},
  booktitle={Proceedings of the IEEE/CVF conference on computer vision and pattern recognition},
  pages={3001--3011},
  year={2022}
}

@inproceedings{PINA,
  title={Non-exemplar Domain Incremental Learning via Cross-Domain Concept Integration},
  author={Wang, Qiang and He, Yuhang and Dong, Songlin and Gao, Xinyuan and Wang, Shaokun and Gong, Yihong},
  booktitle={European Conference on Computer Vision},
  pages={144--162},
  year={2025},
  organization={Springer}
}

@article{FasterRCNN,
  title={Faster r-cnn: Towards real-time object detection with region proposal networks},
  author={Ren, Shaoqing and He, Kaiming and Girshick, Ross and Sun, Jian},
  journal={Advances in neural information processing systems},
  volume={28},
  year={2015}
}

@inproceedings{RandBox,
  title={Random boxes are open-world object detectors},
  author={Wang, Yanghao and Yue, Zhongqi and Hua, Xian-Sheng and Zhang, Hanwang},
  booktitle={Proceedings of the IEEE/CVF international conference on computer vision},
  pages={6233--6243},
  year={2023}
}

@inproceedings{CL-LoRA,
  title={CL-LoRA: Continual Low-Rank Adaptation for Rehearsal-Free Class-Incremental Learning},
  author={He, Jiangpeng and Duan, Zhihao and Zhu, Fengqing},
  booktitle={Proceedings of the Computer Vision and Pattern Recognition Conference},
  pages={30534--30544},
  year={2025}
}

@article{SD-LoRA,
  title={Sd-lora: Scalable decoupled low-rank adaptation for class incremental learning},
  author={Wu, Yichen and Piao, Hongming and Huang, Long-Kai and Wang, Renzhen and Li, Wanhua and Pfister, Hanspeter and Meng, Deyu and Ma, Kede and Wei, Ying},
  journal={arXiv preprint arXiv:2501.13198},
  year={2025}
}

@article{ZiRA,
  title={Zero-shot generalizable incremental learning for vision-language object detection},
  author={Deng, Jieren and Zhang, Haojian and Ding, Kun and Hu, Jianhua and Zhang, Xingxuan and Wang, Yunkuan},
  journal={Advances in Neural Information Processing Systems},
  volume={37},
  pages={136679--136700},
  year={2024}
}
}

\clearpage
\renewcommand{\thesection}{\Alph{section}}
\renewcommand{\thetable}{S\arabic{table}}
\renewcommand{\thefigure}{S\arabic{figure}}
\setcounter{section}{0}
\setcounter{figure}{0}
\setcounter{table}{0}

\maketitlesupplementary

\section{EUMix Mathmematical Formulation}
\label{sec:unknown_mix_supp}

For completeness, we detail the formulation of the Entropy-Aware Unknown Mixing (EUMix) module in this section. For a given query $i$ at task $t$, let
\begin{equation}
\mathbf{z}^{\mathrm{cls}}_i
=
\big[\mathbf{z}^{\mathrm{known}}_i,\, z^{\mathrm{unk}}_i\big]
\end{equation}
denote the raw classification logits over the $|\mathcal{K}^t|$ known classes and the single unknown class, where $\mathbf{z}^{\mathrm{known}}_i \in \mathbb{R}^{|\mathcal{K}^t|}$ and $z^{\mathrm{unk}}_i \in \mathbb{R}$. We write $z^{\mathrm{known}}_{i,c}$ for the logit of known class $c \in \mathcal{K}^t$. In addition, let $z^{\mathrm{obj}}_i$ be the objectness logit produced by the Query-Norm Objectness Adapter for the same query.

We first compute the maximum known-class confidence
\begin{equation}
p^{\mathrm{known},\max}_i
=
\max_{c \in \mathcal{K}^t}
\sigma\big(z^{\mathrm{known}}_{i,c}\big),
\end{equation}
where $\sigma(\cdot)$ denotes the logistic sigmoid. If the model is highly confident in some known class, $p^{\mathrm{known},\max}_i$ is close to $1$, whereas ambiguous or out-of-distribution objects yield lower values. We convert this observation into a calibrated gap $g_i$ that measures how much probability mass is available for the unknown class:
\begin{equation}
g_i = \big(1 - p^{\mathrm{known},\max}_i\big)^{\gamma},
\qquad
\gamma = \mathrm{softplus}(\theta_\gamma),
\end{equation}
where $\theta_\gamma$ is a learned scalar and $\gamma > 0$ acts as a temperature on the gap. When $\gamma > 1$ the gap is sharpened, so that only strongly uncertain predictions yield a significant $g_i$; when $\gamma < 1$ the transition is smoother, which is beneficial if known-class logits are noisy.

We interpret the product of objectness and $g_i$ as an objectness-derived unknown probability:
\begin{equation}
p^{\mathrm{unk}}_{\mathrm{obj},i}
=
\sigma\big(z^{\mathrm{obj}}_i\big)\, g_i,
\end{equation}
which is high exactly when the model believes there is an object at the query location but no known class explains it confidently. In parallel, we convert the learned unknown logit into a probability, allowing a learnable bias $b_{\mathrm{obj}}$ to compensate for the fact that the unknown logit rarely sees positive supervision:
\begin{equation}
p^{\mathrm{unk}}_{\mathrm{cls},i}
=
\sigma\big(z^{\mathrm{unk}}_i + b_{\mathrm{obj}}\big).
\end{equation}

EUMix combines these two estimates through a learnable mixing coefficient
\begin{equation}
\alpha = \sigma(\theta_\alpha),
\end{equation}
where $\theta_\alpha$ is a scalar parameter. The final unknown probability is
\begin{equation}
p^{\mathrm{unk}}_{\mathrm{final},i}
=
\alpha\, p^{\mathrm{unk}}_{\mathrm{cls},i}
+
(1-\alpha)\, p^{\mathrm{unk}}_{\mathrm{obj},i},
\end{equation}
which is then converted back to a logit:
\begin{equation}
z^{\mathrm{unk}}_{\mathrm{final},i}
=
\mathrm{logit}\big(p^{\mathrm{unk}}_{\mathrm{final},i}\big),
\end{equation}
where $\mathrm{logit}$ denotes the inverse of the logistic sigmoid. The mixing weight $\alpha$ is initialised to favour the classifier and is learned end-to-end; if the classifier becomes reliable on unknowns, $\alpha$ naturally increases, but in early tasks or under strong domain shift the model can lean more heavily on the objectness–gap signal.

The final logits fed into the detection loss are then
\begin{equation}
\mathbf{z}^{\mathrm{final}}_i
=
\big[
\mathbf{z}^{\mathrm{known}}_{\mathrm{final},i},\,
z^{\mathrm{unk}}_{\mathrm{final},i}
\big].
\end{equation}
All parameters in this module,
\(
\theta_\gamma, \theta_\alpha, \theta_\lambda, b_{\mathrm{obj}},
\)
are trained jointly with the rest of the network using exactly the same detection loss as the base detector, without any explicit supervision on the unknown category. Their role is purely to reshape the logit space so that unknown evidence coming from objectness and classification uncertainty is translated into calibrated unknown scores. 

\section{Addressing data imbalance across tasks}
\label{sec:data_imbal}
\begin{figure}[ht]
    \centering
    \setlength{\fboxsep}{0pt} 
    \setlength{\fboxrule}{0.8pt} 
    \begin{subfigure}{0.48\columnwidth}
        \centering
        \fbox{\includegraphics[width=\columnwidth]{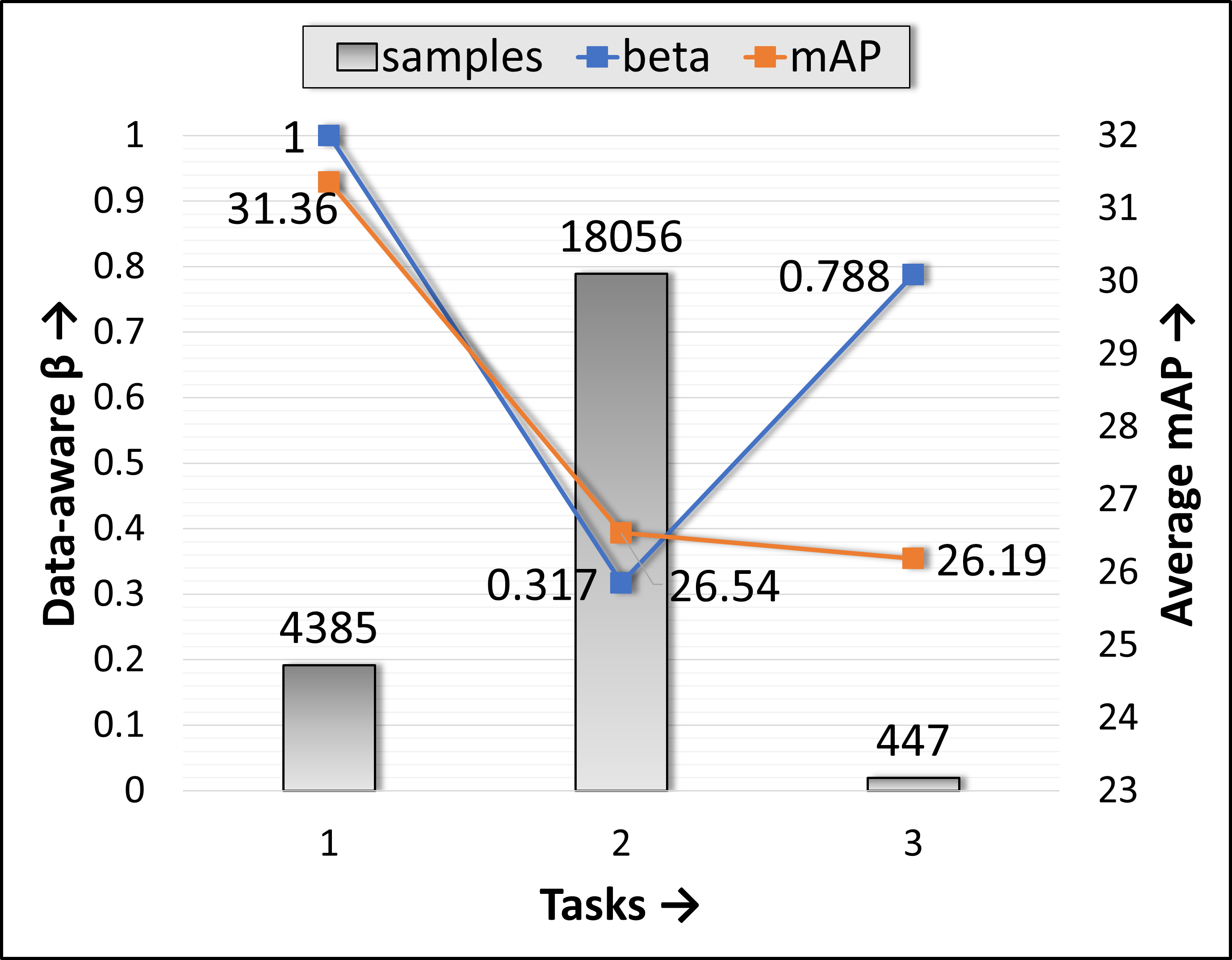}}
        \caption{}
        \label{fig:data_aware_beta}
    \end{subfigure}
    \hfill
    \begin{subfigure}{0.48\columnwidth}
        \centering
        \fbox{\includegraphics[width=\columnwidth]{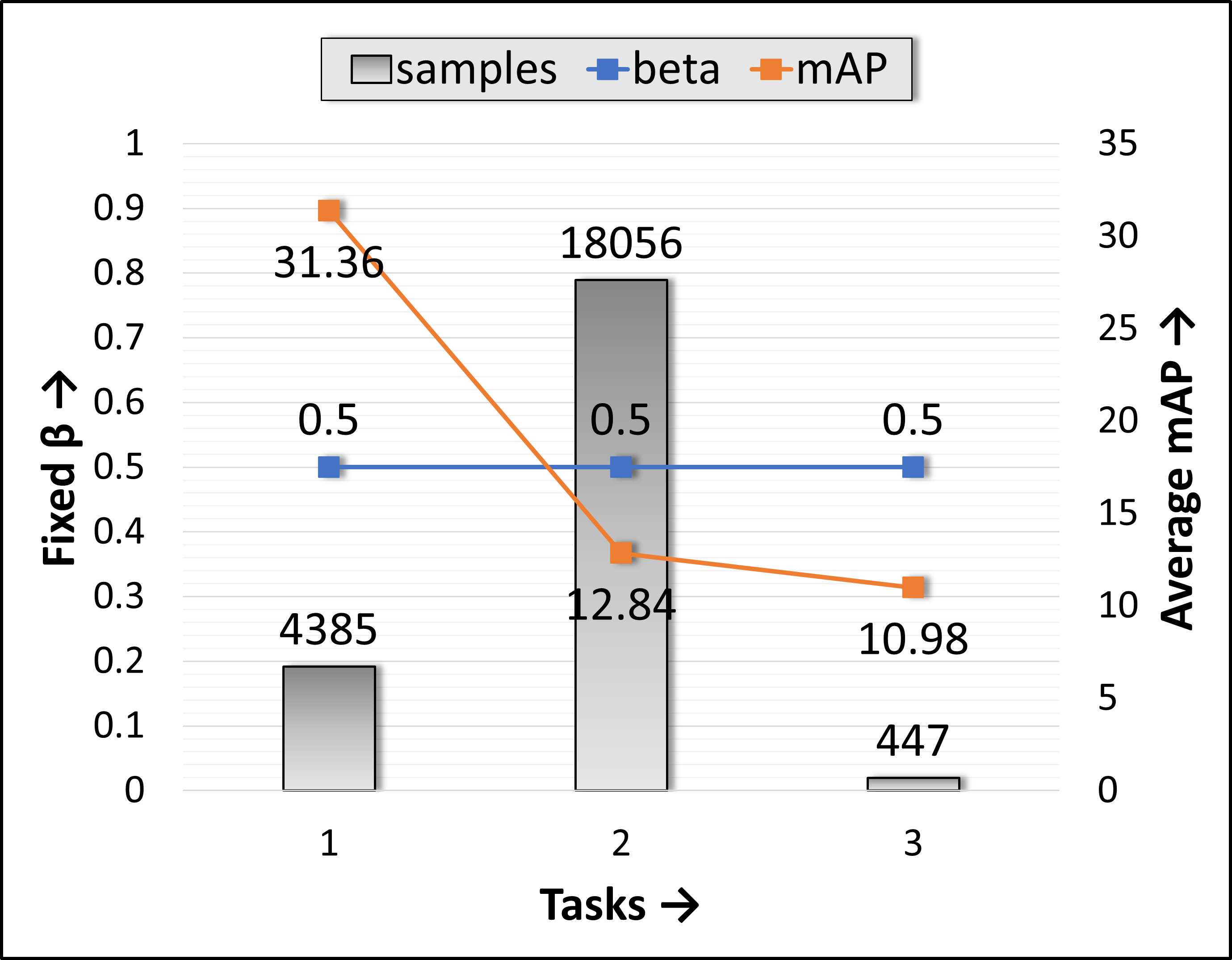}}
        \caption{}
        \label{fig:fixed_beta}
    \end{subfigure}
    \caption{Effect of \subref{fig:data_aware_beta} data-aware vs. \subref{fig:fixed_beta} fixed merging.}
    \vspace{-2mm}
    \label{fig:merging_coeff_effect}
\end{figure}
A distinctive challenge in EWOD is the severe data imbalance across tasks, where different domains and class distributions yield vastly different sample sizes. As illustrated in Table~\ref{tab:dataset_stats_div_weather_multi}, the Diverse Weather benchmark exhibits severe data imbalance across tasks, with Task $\mathcal{T}_2$ (Night Sunny) containing approximately 40 times more training samples than Task $\mathcal{T}_3$ (Night Rainy). This imbalance poses a critical challenge for knowledge consolidation in exemplar-free incremental learning: naively merging task-specific updates into the aggregate adapter can either cause catastrophic forgetting when large tasks dominate, or prevent adaptation when small tasks are under-weighted. In Figure~\ref{fig:data_aware_beta}, our data-aware merging coefficient $\beta_t$ (Eq. 3, main) dynamically adjusts based on the ratio between current and aggregate samples. At Task $\mathcal{T}_1$, $\beta_1 = 1.0$ initialises the aggregate adapter entirely from the first task. At Task $\mathcal{T}_2$, the large training sample count relative to $\mathcal{T}_1$ results in a low $\beta_2 = 0.317$, limiting the influence of this data-rich task to prevent it from overwhelming prior knowledge. Consequently, the model retains strong performance (average mAP: 26.54), while if we compare it to fixed $\beta$ (Figure~\ref{fig:fixed_beta}) case, the uniform weighting causes the data-rich Task $\mathcal{T}_2$ to dominate the aggregate adapter, resulting in severe performance degradation, resulting in severe performance degradation: the average mAP drops from 31.36 to 12.84 at $\mathcal{T}_2$. Hence, the fixed strategy fails to preserve knowledge from data-scarce tasks ($\mathcal{T}_1$, $\mathcal{T}_3$) as their contributions are insufficiently weighted during merging.  The effectiveness of data-aware merging is further validated in Figure~\ref{fig:beta_ablation} (\texttt{Appendix~\ref{sec:hyperparam_sweeps}}), where we ablate different data-aware merging bounds $(\beta_{\min}, \beta_{\max})$ for the merging coefficient $\beta_{t}$. The fixed-$\beta$ baseline $(\beta_{\min} = \beta_{\max} = 0.5)$ achieves the lowest FOGS score (54.04) with catastrophic GSS collapse (0.02), confirming that ignoring data imbalance severely impairs both retention and generalisation. Hence, data-aware merging is crucial for balancing stability and plasticity in EWOD under heterogeneous data distributions.

\section{Query norm across tasks}
\label{sec:query_norm_across_tasks}
\vspace{-2mm}
\begin{figure}[ht]
\begin{center}
\centerline{\includegraphics[width=\columnwidth]{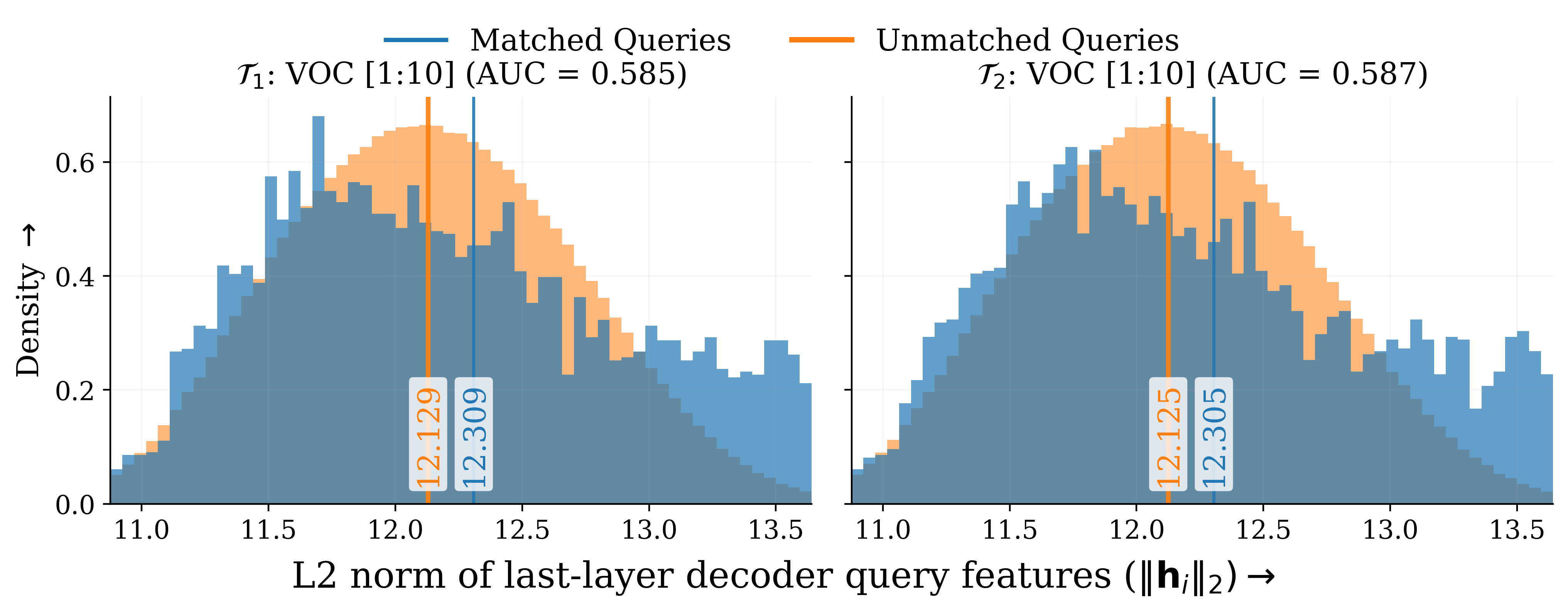}}
\vspace{-3mm}
\caption{Histogram plots of last-layer decoder query norms for VOC [1:10] \(\to\) Clipart [11:18].}
\label{fig:decoder_query_plots}
\end{center}
\vspace{-8mm}
\end{figure}

\noindent To quantitatively validate the norm-objectness hypothesis used by QNorm-Obj (Section 3.4, main), we plot the \(\ell_2\) norm of last-layer decoder query features \(\|\mathbf{h}_i\|_2\) and split queries at inference into \textbf{matched} (assigned to a ground-truth box by Hungarian matching) and \textbf{unmatched} (background). Figure~\ref{fig:decoder_query_plots} shows that matched queries consistently attain higher norms than unmatched ones, both after training on \(\mathcal{T}_1\) (VOC [1:10]: \(12.309 > 12.129\)) and when evaluating the same prior classes after training on \(\mathcal{T}_2\) (VOC [1:10]: \(12.305 > 12.125\)). Moreover, \(\|\mathbf{h}_i\|_2\) maintains stable separation across tasks (ROC-AUC \(=0.585\) in \(\mathcal{T}_1\) and \(0.587\) in \(\mathcal{T}_2\)), supporting that query norms provide a persistent, class-agnostic objectness signal across tasks, as utilized by QNorm-Obj (Section 3.4, main).

\section{EWOD Protocol and Metrics}
\label{sec:eval_protocol_supp}
\begin{table*}[ht]
\centering
\caption{EWOD Training and Evaluation Protocol in Diverse Weather benchmark. \textcolor{blue}{Blue}, \textcolor{orange}{orange}, and \textcolor{olive}{olive} denote classes introduced in $\mathcal{T}_1$, $\mathcal{T}_2$, and $\mathcal{T}_3$, respectively; \textcolor{red}{\sout{red}} indicates classes excluded during training in all tasks but treated as \textcolor{red}{\textbf{\textit{unknown}}} during testing. }
\vspace{-2mm}
\renewcommand{\arraystretch}{1.2}
\setlength{\tabcolsep}{6pt}
\begin{tabular}{c|c|c|c|c}
\hline
\rowcolor{gray!25}
\multicolumn{5}{c}{\textbf{Daytime Sunny [1:3] $\rightarrow$ Night Sunny [4:6]}} \\ \hline
\rowcolor{gray!25}
\textbf{Task} & \textbf{Train Classes} & \textbf{Train Domain} & \textbf{Test Classes} & \textbf{Test Domains} \\ \hline
$\mathcal{T}_1$ &
\begin{tabular}[c]{@{}c@{}}Daytime Sunny [1:3] \\ \textcolor{blue}{bike, bus, car}\end{tabular} &
Daytime Sunny &
\begin{tabular}[c]{@{}c@{}}Daytime Sunny [1:3] \\ \textcolor{blue}{bike, bus, car} + \textcolor{red}{\textbf{\textit{unknown}}}\end{tabular} &
Daytime Sunny \\ \hline
$\mathcal{T}_2$ &
\begin{tabular}[c]{@{}c@{}}Night Sunny [4:6] \\ \textcolor{orange}{motor, person, rider}, \textcolor{red}{\sout{truck}}\end{tabular} &
Night Sunny &
\begin{tabular}[c]{@{}c@{}}Daytime Sunny [1:6] + Night Sunny [1:6] \\ \textcolor{blue}{bike, bus, car} + \textcolor{orange}{motor, person, rider} + \textcolor{red}{\textbf{\textit{unknown}}}\end{tabular} &
\begin{tabular}[c]{@{}c@{}}Daytime Sunny +\\ Night Sunny\end{tabular} \\ 
\hline
\hline
\rowcolor{gray!25}
\multicolumn{5}{c}{\textbf{Daytime Sunny [1:2] $\rightarrow$ Night Sunny [3:4] $\rightarrow$ Night Rainy [5:6]}} \\ \hline
\rowcolor{gray!25}
\textbf{Task} & \textbf{Train Classes} & \textbf{Train Domain} & \textbf{Test Classes} & \textbf{Test Domains} \\ \hline
$\mathcal{T}_1$ &
\begin{tabular}[c]{@{}c@{}}Daytime Sunny [1:2] \\ \textcolor{blue}{bike, bus}\end{tabular} &
Daytime Sunny &
\begin{tabular}[c]{@{}c@{}}Daytime Sunny [1:2] \\ \textcolor{blue}{bike, bus} + \textcolor{red}{\textbf{\textit{unknown}}}\end{tabular} &
Daytime Sunny \\ \hline
$\mathcal{T}_2$ &
\begin{tabular}[c]{@{}c@{}}Night Sunny [3:4] \\ \textcolor{orange}{car, motor}\end{tabular} &
Night Sunny &
\begin{tabular}[c]{@{}c@{}}Daytime Sunny [1:4] + Night Sunny [1:4] \\ \textcolor{blue}{bike, bus} + \textcolor{orange}{car, motor} + \textcolor{red}{\textbf{\textit{unknown}}}\end{tabular} &
\begin{tabular}[c]{@{}c@{}}Daytime Sunny +\\ Night Sunny\end{tabular} \\ \hline
$\mathcal{T}_3$ &
\begin{tabular}[c]{@{}c@{}}Night Rainy [5:6] \\ \textcolor{olive}{person, rider}, \textcolor{red}{\sout{truck}}\end{tabular} &
Night Rainy &
\begin{tabular}[c]{@{}c@{}}Daytime Sunny [1:6] + Night Sunny [1:6] +\\ Night Rainy [1:6] \\ \textcolor{blue}{bike, bus} + \textcolor{orange}{car, motor} + \textcolor{olive}{person, rider} + \textcolor{red}{\textbf{\textit{unknown}}}\end{tabular} &
\begin{tabular}[c]{@{}c@{}}Daytime Sunny +\\ Night Sunny +\\ Night Rainy \end{tabular} \\ \hline
\end{tabular}
\label{tab:eval_protocol}
\vspace{-2mm}
\end{table*}
Table~\ref{tab:eval_protocol} illustrates the proposed EWOD training and evaluation protocol on the Diverse Weather benchmark. EWOD evaluation protocol builds upon the dual-incremental schedule of DuIOD~\cite{DuET}, while integrating open-world constraints from OWOD methods~\cite{ORE,OW-DETR,OWOBJ}. As shown in Table~\ref{tab:eval_protocol}, the protocol spans multiple tasks \(\mathcal{T}_1, \mathcal{T}_2, \ldots, \mathcal{T}_T\), where each task introduces new classes from a new domain.

\vspace{-1em}
\paragraph{Training Protocol.} During training on task \(\mathcal{T}_t\), only the current task's known classes \(\mathcal{K}_t\) receive supervision through bounding box annotations. Crucially, instances of previously learned classes \(\{\mathcal{K}_1 \cup \ldots \cup \mathcal{K}_{t-1}\}\) and truly novel objects remain unlabeled in the training set, mimicking real-world scenarios where exhaustive annotation is impractical. Additionally, certain classes are intentionally withheld from training across all tasks (e.g., \textcolor{red}{\sout{truck}} in Table~\ref{tab:eval_protocol}) to serve as consistent unknown objects throughout the evaluation.

\vspace{-1em}
\paragraph{Evaluation Protocol.} At task \(\mathcal{T}_t\), the detector is evaluated on its ability to: (i) detect all previously learned classes \(\{\mathcal{K}_1 \cup \ldots \cup \mathcal{K}_t\}\) across all encountered domains \(\{\mathcal{D}_1, \ldots, \mathcal{D}_t\}\), (ii) identify unseen objects as \textit{``unknown''}, and (iii) generalize to domain shifts by detecting current task classes across mixed domains. 

To illustrate this protocol, consider the 3-task scenario in Table~\ref{tab:eval_protocol}: \textit{Daytime Sunny [1:2] \(\to\) Night Sunny [3:4] \(\to\) Night Rainy [5:6]}. In \(\mathcal{T}_1\), the model learns \textcolor{blue}{bike, bus} from Daytime Sunny and is evaluated on these classes plus \textcolor{red}{\textbf{\textit{unknown}}} detection in Daytime Sunny. In \(\mathcal{T}_2\), the model learns \textcolor{orange}{car, motor} from Night Sunny (without annotations for \textcolor{blue}{bike, bus}) and is evaluated on \textcolor{blue}{bike, bus} + \textcolor{orange}{car, motor} + \textcolor{red}{\textbf{\textit{unknown}}} across both Daytime Sunny and Night Sunny domains. Finally, in \(\mathcal{T}_3\), the model learns \textcolor{olive}{person, rider} from Night Rainy (excluding \textcolor{red}{\sout{truck}}) and is evaluated on all six learned classes plus unknowns across all three seen domains: Daytime Sunny, Night Sunny, and Night Rainy.

Hence, EWOD protocol ensures that the detector is challenged with: (1) retaining past knowledge without explicit supervision, (2) rejecting unknowns while maintaining precision on known classes, and (3) adapting to domain shifts for both old and new classes.

\vspace{-1em}
\paragraph{FOGS: Forgetting-Openness-Generalisation Score}
As discussed in \texttt{Section 4.2} (main), existing metrics either focus on individual EWOD dimensions or fail to capture their interplay. To capture the coupled failure modes in EWOD and facilitate holistic comparison, we introduce FOGS as the mean of three calibrated sub-scores:
\begin{equation}
    \text{FOGS} = \frac{\text{FSS} + \text{OSS} + \text{GSS}}{3}
\end{equation}
where each sub-score quantifies a distinct dimension of EWOD performance. Higher FOGS indicates better overall performance across forgetting resistance, open-set robustness, and domain generalisation.

\vspace{-1em}
\paragraph{FSS: Forgetting Sub-Score}
As discussed in \texttt{Section 4.2 (main)}, FSS  quantifies the detector's ability to retain performance on previously learned classes across all cumulative tasks. As defined in Eq. 12 (main), FSS measures retention by comparing the performance on previously learned classes at each task with their average initial performance when first introduced.

To illustrate the computation, consider the 3-task scenario from Table~\ref{tab:eval_protocol}: \textit{Daytime Sunny [1:2] \(\to\) Night Sunny [3:4] \(\to\) Night Rainy [5:6]}. Here, we have \(T=3\) tasks, and FSS is computed by averaging retention ratios across tasks \(\mathcal{T}_2\) and \(\mathcal{T}_3\):

\begin{equation}
    \scalebox{0.9}{ $
    \text{FSS} = \frac{1}{2} \left( \frac{\text{mAP}_{\text{prev}}^{\mathcal{T}_2}(\text{bike, bus})}{\text{mAP}_{\text{curr}}^{\mathcal{T}_1}(\text{bike, bus})} + \frac{\text{mAP}_{\text{prev}}^{\mathcal{T}_3}(\text{bike, bus, car, motor})}{\frac{1}{2} \left(\text{mAP}_{\text{curr}}^{\mathcal{T}_1}(\text{bike, bus}) + \text{mAP}_{\text{curr}}^{\mathcal{T}_2}(\text{car, motor})\right)} \right)
    $ }
\end{equation}

The first term measures how well the detector retains \textcolor{blue}{bike, bus} (learned in \(\mathcal{T}_1\) from Daytime Sunny) when evaluated at \(\mathcal{T}_2\) while the second term measures retention of all previously learned classes \textcolor{blue}{bike, bus}, \textcolor{olive}{car, motor} when evaluated at \(\mathcal{T}_3\). A higher FSS indicates better retention of past knowledge, while lower values indicate catastrophic forgetting.

\vspace{-1em}
\paragraph{OSS: Openness Sub-Score}
OSS captures the detector's open-set behaviour by combining three complementary metrics from OWOD literature~\cite{ORE,OW-DETR,OWOBJ}, as defined in Eq. 13 (main). For the 3-task scenario in Table~\ref{tab:eval_protocol}, we compute OSS by averaging the openness scores across all three tasks:
\begin{equation}
    \scalebox{0.99}{$
    \text{OSS} = \frac{1}{3} \sum_{t=1}^{3}
    \left(
    \frac{
    \text{U-Recall}_t + \left(1-\text{WI}_t\right) + \frac{1}{1 + \text{A-OSE}_t / \text{GT}_{\text{unk, t}}}
    }{3}
    \right)
    $}
\end{equation}
Higher OSS values indicate robust unknown detection with minimal impact on known class precision, while lower values suggest open-set collapse.

\vspace{-1em}
\paragraph{GSS: Generalisation Sub-Score}
GSS evaluates the detector's ability to adapt to domain shifts, as defined in Eq. 14 (main). GSS measures how well newly learned classes transfer across the mixed domains encountered so far.

For the 3-task scenario from Table~\ref{tab:eval_protocol}, GSS is computed by averaging the cross-domain performance for tasks \(\mathcal{T}_2\) and \(\mathcal{T}_3\):
\begin{equation}
    \scalebox{0.92}{$
    \text{GSS} = \frac{1}{2} \left( \text{mAP}_{\text{curr}}^{\mathcal{T}_2}(\text{car, motor}) + \text{mAP}_{\text{curr}}^{\mathcal{T}_3}(\text{person, rider}) \right)
    $}
\end{equation}

Here, \(\text{mAP}_{\text{curr}}^{\mathcal{T}_2}(\text{car, motor})\) essentially measures whether the detector can generalise \textcolor{orange}{car, motor} to Daytime Sunny, despite being trained only on Night Sunny. Similarly, \(\text{mAP}_{\text{curr}}^{\mathcal{T}_3}(\text{person, rider})\) measures the performance on classes \textcolor{olive}{person, rider} (learned from Night Rainy) when evaluated across all three Daytime Sunny, Night Sunny and Night Rainy domains, testing cross-domain adaptability for these newly learned classes. Higher GSS reflects stronger domain generalisation, while lower values indicate domain-specific overfitting.

\section{Dataset Statistics}
\label{sec:data_stats}
\begin{table*}[!ht]
    \centering
    \caption{Class-wise distribution on Diverse Weather benchmark: Daytime Sunny [1:2] \(\to\) Night Sunny [3:4] \(\to\) Night Rainy [5:6]; \textcolor{red}{\sout{red}} indicates classes (i.e., \textcolor{red}{\sout{truck}}) excluded during training in all tasks but treated as \textcolor{red}{\textbf{\textit{unknown}}} during testing. Best viewed in colour.}
    \label{tab:dataset_stats_div_weather_multi}
    \vspace{-2mm}
    {\normalsize
    \resizebox{\textwidth}{!}{%
        \begin{tabular}{c|c||ccc|ccc}
        \toprule
        \multirow{2}{*}{\makecell{\textbf{Class} \\ \textbf{ID}}} & \multirow{2}{*}{\makecell{\textbf{Class} \\ \textbf{Name}}} & \multicolumn{3}{c|}{\textbf{Training}} & \multicolumn{3}{c}{\textbf{Testing}} \\
         & & \makecell{\(\mathcal{T}_1:\) \textcolor{cliporange}{Daytime} \\ \textcolor{cliporange}{Sunny}} & \makecell{\(\mathcal{T}_2:\) \textcolor{vocblue}{Night} \\ \textcolor{vocblue}{Sunny}} & \makecell{\(\mathcal{T}_3:\) \textcolor{comicmagenta}{Night} \\ \textcolor{comicmagenta}{Rainy}} & \makecell{\(\mathcal{T}_1:\) \textcolor{cliporange}{Daytime} \\ \textcolor{cliporange}{Sunny}} & \makecell{\(\mathcal{T}_2:\) \textcolor{cliporange}{Daytime Sunny} \\ + \textcolor{vocblue}{Night Sunny}} & \makecell{\(\mathcal{T}_3:\) \textcolor{cliporange}{Daytime Sunny} \\ + \textcolor{vocblue}{Night Sunny} \\ + \textcolor{comicmagenta}{Night Rainy}} \\
        \hline
        1 & bike & \textcolor{cliporange}{\cmark} & & & \textcolor{cliporange}{\cmark} & \textcolor{cliporange}{\cmark}\textcolor{vocblue}{\cmark} & \textcolor{cliporange}{\cmark}\textcolor{vocblue}{\cmark}\textcolor{comicmagenta}{\cmark} \\
        2 & bus & \textcolor{cliporange}{\cmark} & & & \textcolor{cliporange}{\cmark} & \textcolor{cliporange}{\cmark}\textcolor{vocblue}{\cmark} & \textcolor{cliporange}{\cmark}\textcolor{vocblue}{\cmark}\textcolor{comicmagenta}{\cmark} \\
        3 & car & & \textcolor{vocblue}{\cmark} & & & \textcolor{cliporange}{\cmark}\textcolor{vocblue}{\cmark} & \textcolor{cliporange}{\cmark}\textcolor{vocblue}{\cmark}\textcolor{comicmagenta}{\cmark} \\
        4 & motor & & \textcolor{vocblue}{\cmark} & & & \textcolor{cliporange}{\cmark}\textcolor{vocblue}{\cmark} & \textcolor{cliporange}{\cmark}\textcolor{vocblue}{\cmark}\textcolor{comicmagenta}{\cmark} \\
        5 & person & & & \textcolor{comicmagenta}{\cmark} & & & \textcolor{cliporange}{\cmark}\textcolor{vocblue}{\cmark}\textcolor{comicmagenta}{\cmark} \\
        6 & rider & & & \textcolor{comicmagenta}{\cmark} & & & \textcolor{cliporange}{\cmark}\textcolor{vocblue}{\cmark}\textcolor{comicmagenta}{\cmark} \\
        -- & \textcolor{red}{\sout{truck}} & & & & & & \\
        \textcolor{red}{\textbf{7}} & \textcolor{red}{\textbf{\textit{unknown}}} & & & & \textcolor{red}{\cmark} & \textcolor{red}{\cmark} & \textcolor{red}{\cmark} \\
        \hline
        \multicolumn{2}{c||}{\# Classes} & 2 & 2 & 2 & 2+\textcolor{red}{\textbf{1}} & 2+2+\textcolor{red}{\textbf{1}} & 2+2+2+\textcolor{red}{\textbf{1}} \\
        \multicolumn{2}{c||}{\# images} & 4709 & 18459 & 471 & 10219 & 26754 & 29527 \\
        \multicolumn{2}{c||}{\# annotations} & 6772 & 169460 & 1341 & 116785 & 197711 & 364231 \\
        \bottomrule
        \end{tabular}
    }}
\end{table*}
\begin{table*}[!ht]
    \centering
    \caption{Class-wise distribution on three Pascal Series benchmarks: \(\mathcal{T}_1:\) VOC [1:10] \(\to\) \(\mathcal{T}_2:\) \{Clipart [11:18] / Watercolor [11:14] / Comic [11:14]\}. Note that while \textcolor{waterteal}{Watercolor} and \textcolor{comicmagenta}{Comic} classes are listed here as IDs 15--18 for visual continuity, they are strict subsets of the whole label space and are mapped to IDs 11--14 during experiments; \textcolor{red}{\sout{red}} indicates classes (i.e., \textcolor{red}{\sout{dog}}, \textcolor{red}{\sout{person}}) excluded during training in all tasks but treated as \textcolor{red}{\textbf{\textit{unknown}}} during testing. Best viewed in colour.}
    \label{tab:dataset_stats_pascal}
    \vspace{-2mm}
    {\normalsize
    \resizebox{\textwidth}{!}{%
        \begin{tabular}{c|c||c | c c c||c | c c c}
        \toprule
        \multirow{2}{*}{\makecell{\textbf{Class} \\ \textbf{ID}}} & \multirow{2}{*}{\makecell{\textbf{Class} \\ \textbf{Name}}} & \multicolumn{4}{c||}{\textbf{Training}} & \multicolumn{4}{c}{\textbf{Testing}} \\
         & & \(\mathcal{T}_1:\) \textcolor{vocblue}{VOC} & \(\mathcal{T}_2:\) \textcolor{cliporange}{Clipart} & \(\mathcal{T}_2:\) \textcolor{waterteal}{Watercolor} & \(\mathcal{T}_2:\) \textcolor{comicmagenta}{Comic} & \(\mathcal{T}_1:\) \textcolor{vocblue}{VOC} & \(\mathcal{T}_2:\) \textcolor{vocblue}{VOC} + \textcolor{cliporange}{Clipart} & \(\mathcal{T}_2:\) \textcolor{vocblue}{VOC} + \textcolor{waterteal}{Watercolor} & \(\mathcal{T}_2:\) \textcolor{vocblue}{VOC} + \textcolor{comicmagenta}{Comic} \\
        \hline
        1 & aeroplane & \textcolor{vocblue}{\cmark} & & & & \textcolor{vocblue}{\cmark} & \textcolor{vocblue}{\cmark}\textcolor{cliporange}{\cmark} & \textcolor{vocblue}{\cmark} & \textcolor{vocblue}{\cmark} \\
        2 & boat & \textcolor{vocblue}{\cmark} & & & & \textcolor{vocblue}{\cmark} & \textcolor{vocblue}{\cmark}\textcolor{cliporange}{\cmark} & \textcolor{vocblue}{\cmark} & \textcolor{vocblue}{\cmark} \\
        3 & bottle & \textcolor{vocblue}{\cmark} & & & & \textcolor{vocblue}{\cmark} & \textcolor{vocblue}{\cmark}\textcolor{cliporange}{\cmark} & \textcolor{vocblue}{\cmark} & \textcolor{vocblue}{\cmark} \\
        4 & bus & \textcolor{vocblue}{\cmark} & & & & \textcolor{vocblue}{\cmark} & \textcolor{vocblue}{\cmark}\textcolor{cliporange}{\cmark} & \textcolor{vocblue}{\cmark} & \textcolor{vocblue}{\cmark} \\
        5 & chair & \textcolor{vocblue}{\cmark} & & & & \textcolor{vocblue}{\cmark} & \textcolor{vocblue}{\cmark}\textcolor{cliporange}{\cmark} & \textcolor{vocblue}{\cmark} & \textcolor{vocblue}{\cmark} \\
        6 & cow & \textcolor{vocblue}{\cmark} & & & & \textcolor{vocblue}{\cmark} & \textcolor{vocblue}{\cmark}\textcolor{cliporange}{\cmark} & \textcolor{vocblue}{\cmark} & \textcolor{vocblue}{\cmark} \\
        7 & diningtable & \textcolor{vocblue}{\cmark} & & & & \textcolor{vocblue}{\cmark} & \textcolor{vocblue}{\cmark}\textcolor{cliporange}{\cmark} & \textcolor{vocblue}{\cmark} & \textcolor{vocblue}{\cmark} \\
        8 & horse & \textcolor{vocblue}{\cmark} & & & & \textcolor{vocblue}{\cmark} & \textcolor{vocblue}{\cmark}\textcolor{cliporange}{\cmark} & \textcolor{vocblue}{\cmark} & \textcolor{vocblue}{\cmark} \\
        9 & motorbike & \textcolor{vocblue}{\cmark} & & & & \textcolor{vocblue}{\cmark} & \textcolor{vocblue}{\cmark}\textcolor{cliporange}{\cmark} & \textcolor{vocblue}{\cmark} & \textcolor{vocblue}{\cmark} \\
        10 & pottedplant & \textcolor{vocblue}{\cmark} & & & & \textcolor{vocblue}{\cmark} & \textcolor{vocblue}{\cmark}\textcolor{cliporange}{\cmark} & \textcolor{vocblue}{\cmark} & \textcolor{vocblue}{\cmark} \\
        11 & sheep & & \textcolor{cliporange}{\cmark} & & & & \textcolor{vocblue}{\cmark}\textcolor{cliporange}{\cmark} & & \\
        12 & sofa & & \textcolor{cliporange}{\cmark} & & & & \textcolor{vocblue}{\cmark}\textcolor{cliporange}{\cmark} & & \\
        13 & train & & \textcolor{cliporange}{\cmark} & & & & \textcolor{vocblue}{\cmark}\textcolor{cliporange}{\cmark} & & \\
        14 & tvmonitor & & \textcolor{cliporange}{\cmark} & & & & \textcolor{vocblue}{\cmark}\textcolor{cliporange}{\cmark} & & \\
        15 & bicycle & & \textcolor{cliporange}{\cmark} & \textcolor{waterteal}{\cmark} & \textcolor{comicmagenta}{\cmark} & & \textcolor{vocblue}{\cmark}\textcolor{cliporange}{\cmark} & \textcolor{vocblue}{\cmark}\textcolor{waterteal}{\cmark} & \textcolor{vocblue}{\cmark}\textcolor{comicmagenta}{\cmark} \\
        16 & bird & & \textcolor{cliporange}{\cmark} & \textcolor{waterteal}{\cmark} & \textcolor{comicmagenta}{\cmark} & & \textcolor{vocblue}{\cmark}\textcolor{cliporange}{\cmark} & \textcolor{vocblue}{\cmark}\textcolor{waterteal}{\cmark} & \textcolor{vocblue}{\cmark}\textcolor{comicmagenta}{\cmark} \\
        17 & car & & \textcolor{cliporange}{\cmark} & \textcolor{waterteal}{\cmark} & \textcolor{comicmagenta}{\cmark} & & \textcolor{vocblue}{\cmark}\textcolor{cliporange}{\cmark} & \textcolor{vocblue}{\cmark}\textcolor{waterteal}{\cmark} & \textcolor{vocblue}{\cmark}\textcolor{comicmagenta}{\cmark} \\
        18 & cat & & \textcolor{cliporange}{\cmark} & \textcolor{waterteal}{\cmark} & \textcolor{comicmagenta}{\cmark} & & \textcolor{vocblue}{\cmark}\textcolor{cliporange}{\cmark} & \textcolor{vocblue}{\cmark}\textcolor{waterteal}{\cmark} & \textcolor{vocblue}{\cmark}\textcolor{comicmagenta}{\cmark} \\
        -- & \textcolor{red}{\sout{dog}} & & & & & & & & \\
        -- & \textcolor{red}{\sout{person}} & & & & & & & & \\
        \textcolor{red}{\textbf{19}} & \textcolor{red}{\textbf{\textit{unknown}}} & & & & & \textcolor{red}{\cmark} & \textcolor{red}{\cmark} & \textcolor{red}{\cmark} & \textcolor{red}{\cmark} \\
        \hline
        \multicolumn{2}{c||}{\# Classes} & 10 & 8 & 4 & 4 & $10+\textcolor{red}{\textbf{1}}$ & $10+8+\textcolor{red}{\textbf{1}}$ & $10+4+\textcolor{red}{\textbf{1}}$ & $10+4+\textcolor{red}{\textbf{1}}$ \\
        \multicolumn{2}{c||}{\# images} & 8909 & 165 & 1072 & 1150 & 6041 & 7774 & 8657 & 16630 \\
        \multicolumn{2}{c||}{\# annotations} & 16181 & 270 & 1661 & 3214 & 14976 & 16502 & 8719 & 18151 \\
        \bottomrule
        \end{tabular}
    }}
\end{table*}
As mentioned in \texttt{Section 4.1 (main)}, we evaluate our approach on two dataset series that support evolving open-world object detection across domain-incremental tasks. Here, we provide detailed statistics for both benchmarks.

\vspace{-1em}
\paragraph{Diverse Weather benchmark} Table~\ref{tab:dataset_stats_div_weather_multi} presents the class-wise distribution across three weather conditions: Daytime Sunny, Daytime Foggy, and Night Rainy, sourced from BDD-100k~\cite{BDD100k}, FoggyCityscapes~\cite{FoggyCityscapes}, and Adverse-Weather~\cite{AdverseWeather}. The benchmark follows a 2+2+2 class split across three incremental tasks $\mathcal{T}_1$, $\mathcal{T}_2$, and $\mathcal{T}_3$. Each task introduces two new object categories while encountering data from a new weather domain. The testing phase at each task evaluates on the union of all domains seen up to that point, including instances of the \textcolor{red}{\sout{truck}} class treated as \textcolor{red}{\textbf{\textit{unknown}}} in all tasks to simulate open-world scenarios. The dataset exhibits significant domain shift across weather conditions, with Daytime Foggy providing substantially more annotations compared to the other two domains, reflecting realistic data availability patterns across different environmental conditions.

\vspace{-1em}
\paragraph{Pascal Series benchmark} Table~\ref{tab:dataset_stats_pascal} illustrates the class distribution across four visually distinct domains: Pascal VOC~\cite{PASCAL_VOC}, Clipart, Watercolor, and Comic~\cite{PASCAL_Series} for three different Pascal Series benchmarks: VOC [1:10] \(\to\) Clipart [11:18], VOC [1:10] \(\to\) Watercolor [11:14], and VOC [1:10] \(\to\) Comic [11:14]. Each benchmark follows a two-task incremental learning setup where \(\mathcal{T}_1\) trains on VOC with 10 base classes, followed by \(\mathcal{T}_2\) introducing a different domain variant. The Clipart configuration introduces 8 new classes following class split of 10 + 8, while Watercolor and Comic configurations each introduce 4 classes following class split of 10 + 4. Notably, Watercolor and Comic contain only 6 classes in total, which are strict subsets of the 20-class label space shared by VOC and Clipart. Consequently, these domains lack annotations for the 14 classes present in VOC and Clipart. Similar to the Diverse Weather Series, the \textcolor{red}{\sout{dog}} and \textcolor{red}{\sout{person}} classes are excluded during training across all tasks and treated as \textcolor{red}{\textit{unknown}} during testing. The testing phase for each \(\mathcal{T}_2\) configuration evaluates on the union of VOC and the respective domain, with a significant data imbalance as reflected in the counts of images and annotations.

\section{Implementation Details}
\label{sec:implementation_details}
We instantiate EW-DETR on top of DETR-based detectors (RF-DETR-N~\cite{RF-DETR} and Deformable DETR~\cite{D-DETR}), retaining the original detector pipelines and losses unchanged. As discussed in \texttt{Section 3 (main)}, for each task, the backbone and transformer encoder–decoder layers are kept frozen, along with Aggregate LoRA adapters. For Incremental LoRA Adapters, we set rank r = 16, while for data-aware merging coefficient calculation, we keep \((\beta_{min}, \beta_{max})\) = (0.2, 0.8), as detailed in \texttt{Appendix~\ref{sec:hyperparam_sweeps}}. All methods are based on DINO-pretrained ResNet-50 backbones~\cite{DINO, RN50}, while the recent RF-DETR-N supports DINOv2-S~\cite{RF-DETR, DINOv2}. To ensure a fair comparison across baselines, we keep each method’s trainable components consistent with their original implementations; only the dataloaders and training loops are adapted to meet the exemplar-free EWOD requirements. For~\cite{DuET}, we additionally incorporate Energy Based Unknown Identification (EBUI) from \cite{ORE} to handle unknowns; hence, we see non-zero OSS for~\cite{DuET} in Figure~\ref{fig:comparison_with_all}. For all methods in each task, the detector is trained for 100 epochs using the AdamW optimiser with an initial learning rate of 1e-4, weight decay of 1e-4, and batch size of 16 on a single NVIDIA H100 80GB card. All methods follow the evaluation protocol detailed in \texttt{Appendix~\ref{sec:eval_protocol_supp}} for a fair comparison. 

\section{Additional Results \& Ablations}
\label{sec:ablations_supp}
\subsection{Complexity Analysis}
\label{sec:complexity_analysis}
\vspace{-2mm}
\begin{table}[!ht]
    \centering
    \caption{Computational complexity analysis of various methods evaluated on Pascal Series (VOC [1:10] \(\to\) Clipart [11:18]) benchmark, all evaluated on a single NVIDIA H100 80GB card.}
    \label{tab:complexity_analysis}
    \vspace{-2mm}
    \renewcommand{\arraystretch}{1.2}
    \resizebox{\columnwidth}{!}{%
    \begin{tabular}{cc|cccc}
    \toprule
        \multirow{2}{*}{\textbf{Method}}
         & \multirow{2}{*}{\makecell{\textbf{Underlying} \\ \textbf{Detector}}} & \multirow{2}{*}{\makecell{\textbf{Trainable} \\ \textbf{Params (M)}}} & \multirow{2}{*}{\textbf{GFLOPs}} & \multirow{2}{*}{\makecell{\textbf{Avg. Inference} \\ \textbf{Speed (ms)}}} & \multirow{2}{*}{\makecell{\textbf{Avg. Memory} \\ \textbf{Footprint (GB)}}} 
        \\
        & & & & \\
        \hline
        ORE~\cite{ORE} & Faster RCNN~\cite{FasterRCNN} & 32.96 & 1665.03 & 60.41 & 3.75 \\
        OW-DETR~\cite{OW-DETR} & D-DETR~\cite{D-DETR} & 24.22 & 157.71 & 59.82 & 1.33 \\
        PROB~\cite{PROB} & D-DETR~\cite{D-DETR} & 23.99 & 135.31 & 97.71 & 1.32 \\
        CAT~\cite{CAT} & D-DETR~\cite{D-DETR} & 24.25 & 135.34 & 112.54 & 1.35 \\
        ORTH~\cite{ORTH} & RandBox~\cite{RandBox} & 105.9 & 2073.57 & 1151.19 & 1.26 \\
        DuET~\cite{DuET} & D-DETR~\cite{D-DETR} & 24.22 & 135.3 & 150.61 & 1.41 \\
        OWOBJ~\cite{OWOBJ} & D-DETR~\cite{D-DETR} & 23.99 & 135.3 & 80.85 & 1.32 \\
        \rowcolor{lightgrayblue} \textbf{EW-DETR} & D-DETR~\cite{D-DETR} & 0.46 & 171.23 & 131.92 & 1.55 \\
        \rowcolor{lightgrayblue} \textbf{EW-DETR} & RF-DETR~\cite{RF-DETR} & 1.8 & 32.22 & 57.38 & 0.32 \\
    \bottomrule
    \end{tabular}
    }
\end{table}
\vspace{-2mm}
Table~\ref{tab:complexity_analysis} presents a comprehensive computational complexity analysis of EW-DETR against recent methods on the Pascal Series benchmark. Since Incremental LoRA adapters (\texttt{Section 3.3, main}) freeze the base model and transformer encoder-decoder weights, EW-DETR achieves a drastic reduction in trainable parameters: only \textbf{0.46M} when built upon Deformable DETR~\cite{D-DETR} and \textbf{1.8M} when built upon RF-DETR~\cite{RF-DETR}, compared to 23.99M--105.9M for other methods. This represents a \textbf{98.1\%} parameter reduction relative to standard Deformable DETR-based approaches, making EW-DETR highly suitable for resource-constrained deployment scenarios. Moreover, under EWOD, Deformable DETR-based methods with the standard six encoder–decoder layers consistently collapse to near-zero mAP, while a single encoder–decoder layer yields stable results for all; thus, we adopt the single-layer setting throughout. This reduced the number of trainable parameters; hence, we observe \(\sim\)24M trainable parameters for D-DETR-based methods in Table~\ref{tab:complexity_analysis}.

Regarding comparison of FLOPs, the D-DETR variant requires 171.23 GFLOPs, which is moderately higher than other D-DETR-based methods (135.3--157.71 GFLOPs) due to the QNorm-Obj and EUMix modules, yet substantially lower than ORE~\cite{ORE} (1665.03 GFLOPs) and ORTH~\cite{ORTH} (2073.57 GFLOPs). However, the RF-DETR variant achieves remarkable efficiency with only 32.22 GFLOPs. Inference speed remains practical, with EW-DETR (D-DETR) averaging 131.92 ms per image, comparable to other D-DETR methods, while EW-DETR (RF-DETR) achieves 57.38 ms, making it among the fastest approaches evaluated. Following~\cite{RF-DETR}, we use PyTorch's \href{https://github.com/pytorch/pytorch/blob/baee623691a38433d10843d5bb9bc0ef6a0feeef/torch/utils/flop_counter.py#L596}{\texttt{FlopCounterMode}} to calculate FLOPs for all methods.

Memory footprint analysis reveals that EW-DETR (D-DETR) requires 1.55 GB, slightly higher than other D-DETR methods (1.32--1.41 GB) due to the dual LoRA adapter architecture maintaining both aggregate and task-specific adapters during training. However, EW-DETR (RF-DETR) depicts a memory efficiency at only 0.32 GB, significantly lower than all compared methods. Hence, overall complexity analysis demonstrates that the EW-DETR framework is suitable for real and dynamic evolving world scenarios where continuous adaptation, storage constraints, and deployment efficiency are required.
\vspace{-2mm}
\subsection{Ablation analysis for random task sequences}
\label{sec:random_order}
\vspace{-4mm}
\begin{table}[h]
    \centering
    \caption{Sensitivity analysis for random task permutations on Diverse Weather benchmark.}
    \label{tab:random_order}
    \vspace{-2mm}
    \renewcommand{\arraystretch}{1.2}
    \resizebox{\columnwidth}{!}{%
    \begin{tabular}{ccc|cccc} 
        \toprule
        \textbf{$\mathcal{T}_1$} & \textbf{$\mathcal{T}_2$} & \textbf{$\mathcal{T}_3$} & \textbf{FSS} & \textbf{OSS} & \textbf{GSS} & \textbf{FOGS} \\
        \hline
        Daytime Sunny [5:6] & Night Sunny [1:2] & Night Rainy [3:4] & 73.30 & 73.45 & 24.62 & 57.12 \\
        Daytime Sunny [3:4] & Night Sunny [5:6] & Night Rainy [1:2] & 75.03 & 71.31 & 17.94 & 54.76 \\
        Daytime Sunny [1:2] & Night Sunny [3:4] & Night Rainy [5:6] & 73.63 & 73.43 & 18.68 & 55.25 \\
        Night Rainy [1:2] & Daytime Sunny [3:4] & Night Sunny [5:6] & 71.79 & 72.50 & 28.67 & 57.65 \\
        Night Sunny [1:2] & Night Rainy [3:4] & Daytime Sunny [5:6] & 71.93 & 70.69 & 21.25 & 54.62 \\
        Daytime Sunny [7,2] & Night Sunny [3,4] & Night Rainy [5,6] & 73.65 & 71.71 & 14.96 & 53.44 \\
        Daytime Sunny [1,7] & Night Sunny [3,4] & Night Rainy [5,6] & 74.94 & 69.84 & 14.74 & 53.17 \\
        Daytime Sunny [1,2] & Night Sunny [7,4] & Night Rainy [5,6] & 73.36 & 70.2 & 17.72 & 53.76 \\
        \hline
        \multicolumn{3}{c|}{\textbf{Standard Deviation across tasks:}} & \textbf{1.19} & \textbf{1.39} & \textbf{4.81} & \textbf{1.65} \\
        \bottomrule
    \end{tabular}
    }
    \vspace{-2mm}
\end{table}
\begin{table*}[ht]
    \centering
    \caption{Analysis of unknown object confusion on Pascal Series benchmark: VOC [1:10] $\to$ Clipart [11:18]. The table compares all methods using unknown class confusion metrics, including U-Recall, WI and A-OSE. Best results per column in \textbf{bold}, second-best \underline{\textit{underlined}}.}
    \label{tab:E1_WI_AOSE}
    \vspace{-2mm}
    \renewcommand{\arraystretch}{1.2}
    \resizebox{\textwidth}{!}{%
        \begin{tabular}{l||ccc|ccc|cccc}
            \toprule
            \multirow{2}{*}{\textbf{Method}} &
            \multicolumn{3}{c|}{\textbf{\(\mathcal{T}_1\): VOC [1:10]}} &
            \multicolumn{3}{c|}{\textbf{\(\mathcal{T}_2\): Clipart [11:18]}} &
            \multicolumn{4}{c}{\textbf{Metrics}} \\
            
            &  
            \cellcolor{verylightyellow}\textbf{U-Recall ($\uparrow$)} &
            \cellcolor{verylightyellow}\textbf{WI ($\downarrow$)} &
            \cellcolor{verylightyellow}\textbf{A-OSE ($\downarrow$)} &
            \cellcolor{verylightyellow}\textbf{U-Recall ($\uparrow$)} &
            \cellcolor{verylightyellow}\textbf{WI ($\downarrow$)} &
            \cellcolor{verylightyellow}\textbf{A-OSE ($\downarrow$)} &
            \cellcolor{verylightorange}\textbf{FSS ($\uparrow$)} &
            \cellcolor{verylightorange}\textbf{OSS ($\uparrow$)} &
            \cellcolor{verylightorange}\textbf{GSS ($\uparrow$)} &
            \cellcolor{verylightorange}\textbf{FOGS ($\uparrow$)} \\ \hline \hline
            
            ORE-EBUI~\cite{ORE} & 9.84 & 0.0985 & 12916 & 6.97 & 0.0592 & \textbf{852} & 0 & 55.48 & \underline{\textit{11.37}} & 22.28 \\
            
            OW-DETR~\cite{OW-DETR} & 16.25 & 0.0851 & 55228 & 8.07 & 0.0644 & 28689 & 11.42 & 40.47 & 7.96 & 19.95 \\
            
            PROB~\cite{PROB} & 52.73 & 0.1603 & \textbf{12700} & \underline{\textit{46.27}} & 0.0216 & \underline{\textit{1529}} & 0 & \underline{\textit{67.58}} & 0.27 & 22.62 \\
            
            CAT~\cite{CAT} & 20.23 & \underline{\textit{0.083}} & 57094 & 8.1 & 0.0566 & 28045 & 29.52 & 41.29 & 8.05 & 26.29 \\
            
            ORTH~\cite{ORTH} & \underline{\textit{63.92}} & 0.757 & 32519 & 41.61 & \underline{\textit{0.0213}} & 5160 & 5.83 & 51.06 & \textbf{32.44} & 29.78 \\
            
            DuET~\cite{DuET} & 0 & 0.0714 & 80714 & 0 & 0.0622 & 35214 & 41.05 & 35.49 & 1.46 & 26 \\
            
            OWOBJ~\cite{OWOBJ} & 45.93 & 0.155 & 14947 & 35.72 & 0.0383 & 3898 & 0 & 60.73 & 0.51 & 20.41 \\
            
            \rowcolor{lightgrayblue} \textbf{EW-DETR}$_{\text{ D-DETR}}$ & 44.98 & 0.1382 & 20236 & 41.45 & 0.1096 & 2086 & \underline{\textit{64.86}} & 61.67 & 7.92 & \underline{\textit{44.82}} \\
            
            \rowcolor{lightgrayblue} \textbf{EW-DETR}$_{\text{ RF-DETR}}$ & \textbf{77.35} & \textbf{0.012} & \underline{\textit{12773}} & \textbf{78.23} & \textbf{0.0038} & 2251 & \textbf{96.19} & \textbf{78.62} & 8.42 & \textbf{61.08} \\
            \bottomrule
        \end{tabular}%
    }
\end{table*}

\noindent In real-world, evolving environments, the order in which new classes and domains appear can vary unpredictably. To evaluate the robustness of EW-DETR under such conditions, we conducted experiments with different task permutations on the Diverse Weather benchmark. Following~\cite{DuET}, we randomly shuffled both the class assignments and the domain order across five distinct configurations. As shown in Table~\ref{tab:random_order}, EW-DETR delivers consistent performance across all permutations. FSS remains highly stable, with a standard deviation of only \textbf{1.19}, indicating that the proposed Incremental LoRA Adapters effectively preserve previously learned knowledge regardless of task order. Similarly, OSS exhibits very low standard deviation (\textbf{1.11}), demonstrating that the QNorm-Obj and EUMix modules maintain strong unknown-detection capabilities across different sequences. However, GSS shows comparatively larger variation, which could be attributed to significant data imbalance across domains: because GSS measures how well current-task classes generalise to all previously seen domains, configurations in which data-rich domains appear in later tasks naturally achieve higher GSS. Conversely, when data-scarce domains appear in later tasks,  GSS declines noticeably. Despite these variations in individual sub-scores, the overall FOGS metric remains stable with a standard deviation of \textbf{1.26}, confirming that EW-DETR maintains consistent holistic performance across the three critical EWOD dimensions irrespective of task ordering. Moreover, we test \textbf{unknown-class sensitivity} by changing which category is withheld as the stationary unknown prior (last three rows of Table~\ref{tab:random_order}). The performance remains stable under this change: OSS varies only modestly (69.84--71.71), and FOGS remains tightly bounded (53.17--53.76), indicating that QNorm-Obj and EUMix do not rely on a particular choice of unknown class to maintain open-set robustness.

\subsection{Ablation analysis for key hyperparameters}
\label{sec:hyperparam_sweeps}
\begin{figure*}[ht]
    \centering
    \setlength{\fboxsep}{0pt} 
    \setlength{\fboxrule}{0.8pt} 
    \begin{subfigure}[b]{0.48\textwidth}
        \centering
        \fbox{\includegraphics[width=\columnwidth, height=4.5cm]{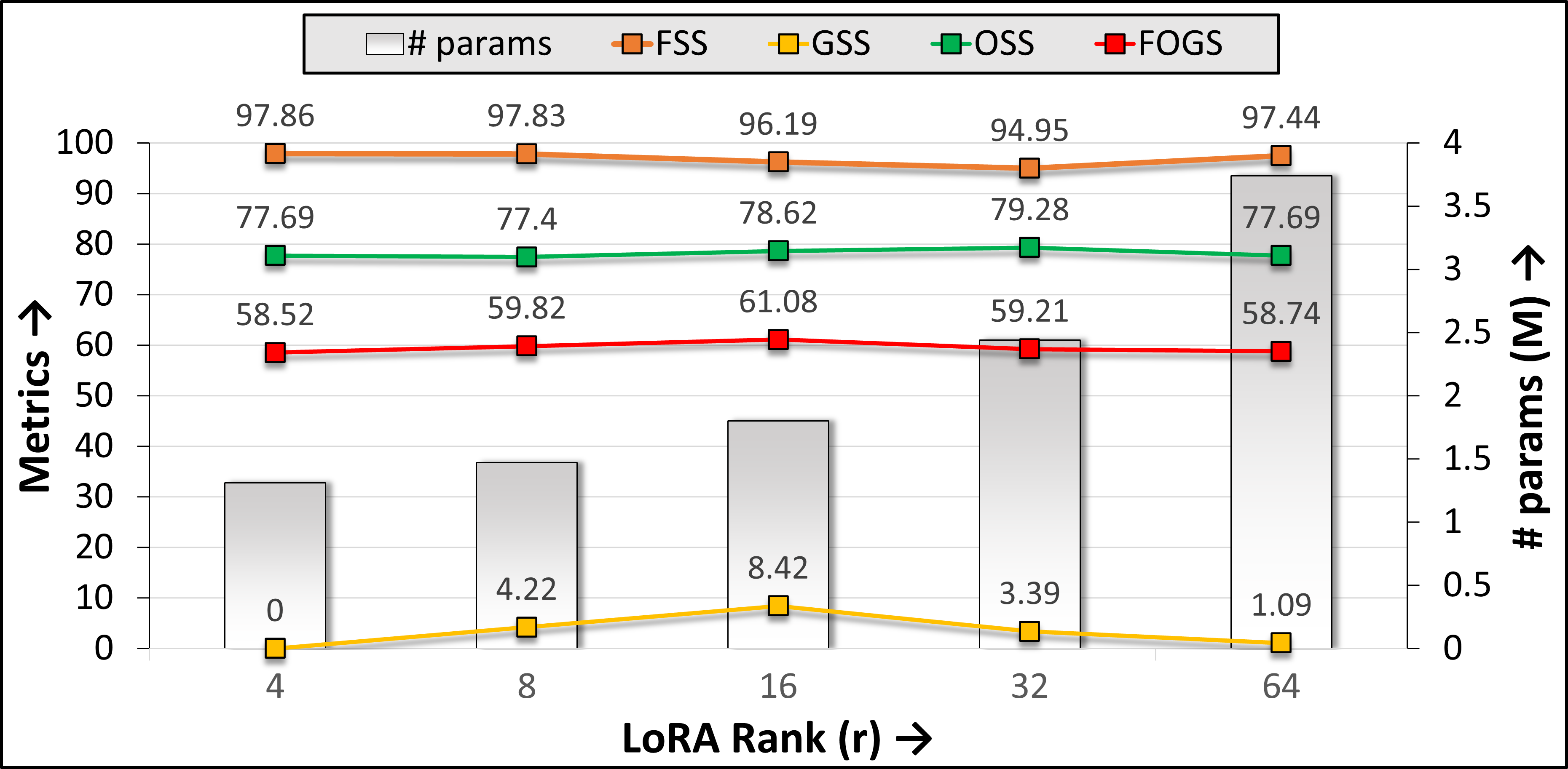}}
        \caption{Sensitivity to LoRA rank \(r\)}
        \label{fig:lora_ablation}
    \end{subfigure}
    \hfill
    \begin{subfigure}[b]{0.48\textwidth}
        \centering
        \fbox{\includegraphics[width=\columnwidth, height=4.5cm]{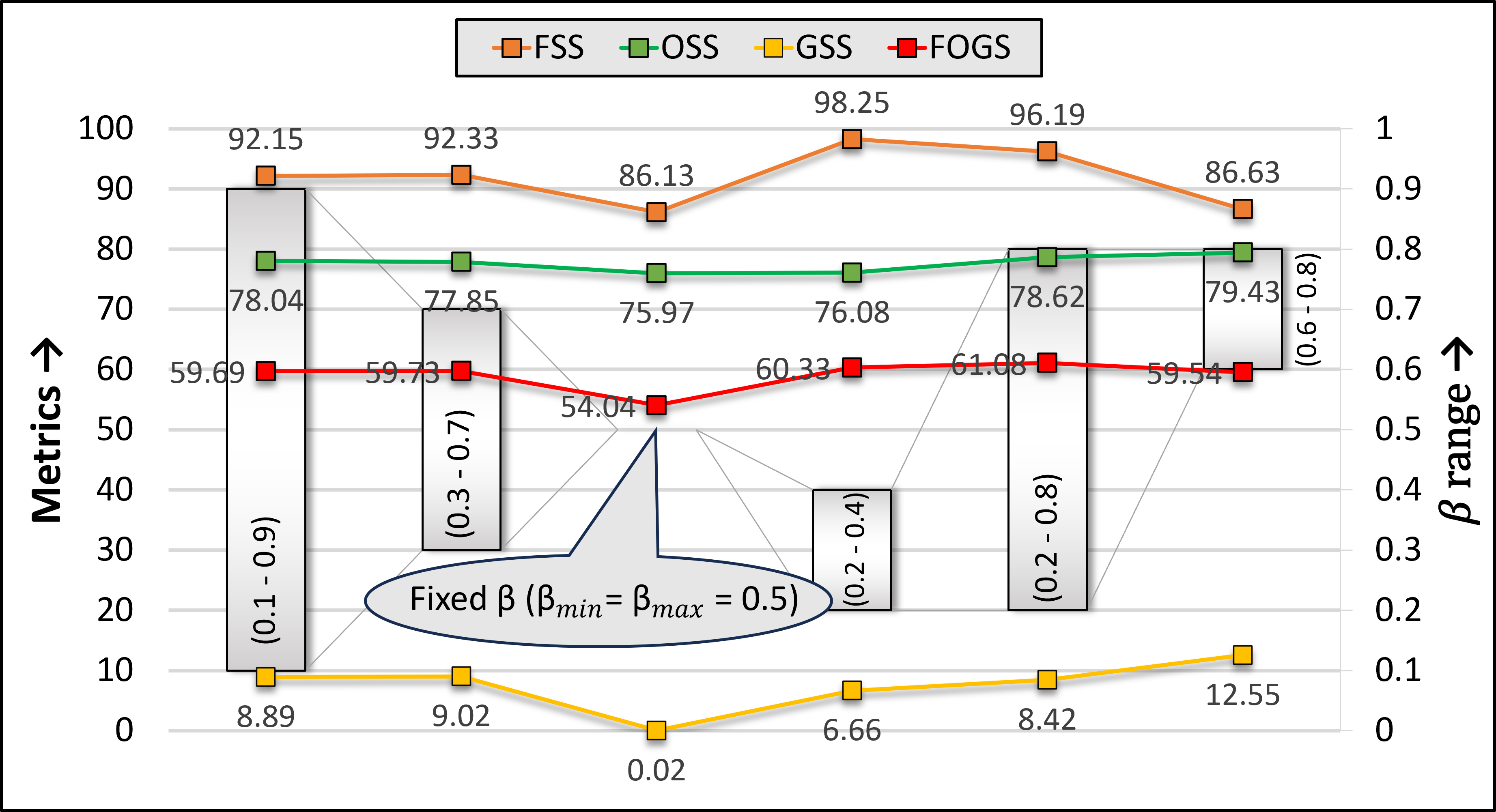}}
        \caption{Sensitivity to data-aware merging bounds \((\beta_{\min},\beta_{\max})\)}
        \label{fig:beta_ablation}
    \end{subfigure}

    \caption{\textbf{Ablation analysis for key hyperparameters in the EW-DETR framework.} 
    \subref{fig:lora_ablation} Sensitivity to LoRA rank \(r\). 
    \subref{fig:beta_ablation} Sensitivity to data-aware merging bounds \((\beta_{\min},\beta_{\max})\).}
    \label{fig:hyperparam_sensitivity}
    \vspace{-4mm}
\end{figure*}

\paragraph{LoRA rank (\(r\)).} 
Figure~\ref{fig:lora_ablation} demonstrates the sensitivity of EW-DETR to LoRA rank \(r\), which governs the capacity of the Incremental LoRA Adapters. FSS remains remarkably stable across all ranks (94.95--97.86). OSS similarly exhibits robustness (77.4--79.28), indicating that the QNorm-Obj and EUMix operate reliably regardless of adapter capacity. However, GSS varies more significantly, peaking at \(r=16\) (8.42) before declining at higher ranks, suggesting that excessive capacity may lead to domain-specific overfitting. The overall FOGS metric peaks at \(r=16\) (61.08) while maintaining a manageable parameter count (1.8 M), representing the optimal trade-off between model capacity and generalisation. Hence, we select \(r=16\) as the default (\texttt{Appendix~\ref{sec:implementation_details}}) for EW-DETR.

\paragraph{Data-aware merging bounds \((\beta_{\min},\beta_{\max})\).}
Figure~\ref{fig:beta_ablation} illustrates how the data-aware merging coefficient bounds impact the stability-plasticity trade-off during LoRA adapter fusion. The fixed-\(\beta\) baseline (\(\beta_{\min}=\beta_{\max}=0.5\)), which ignores task-specific data imbalance, exhibits severe GSS collapse (0.02) and the lowest FOGS (54.04), demonstrating that uniform merging fails under heterogeneous data distributions. Configuration biased towards stability \((0.2,0.4)\) achieve the highest FSS (98.25) by limiting the influence of new tasks, effectively preventing forgetting but at the cost of reduced plasticity (GSS: 6.66). Conversely, plasticity-biased configurations \((0.6,0.8)\) sacrifice some retention (FSS: 86.63) but enable better adaptation to current domains (GSS: 12.55). Overall, \((\beta_{\min},\beta_{\max})=(0.2,0.8)\) emerges as the optimal choice for EWOD, and hence is adopted as the default setting (\texttt{Appendix~\ref{sec:implementation_details}}), achieving a balanced trade-off with FOGS score of 61.08.

\subsection{Comprehensive Results}
\label{sec:quant_results_supp}
\begin{figure*}[ht]
\begin{center}
\centerline{\includegraphics[width=\textwidth]{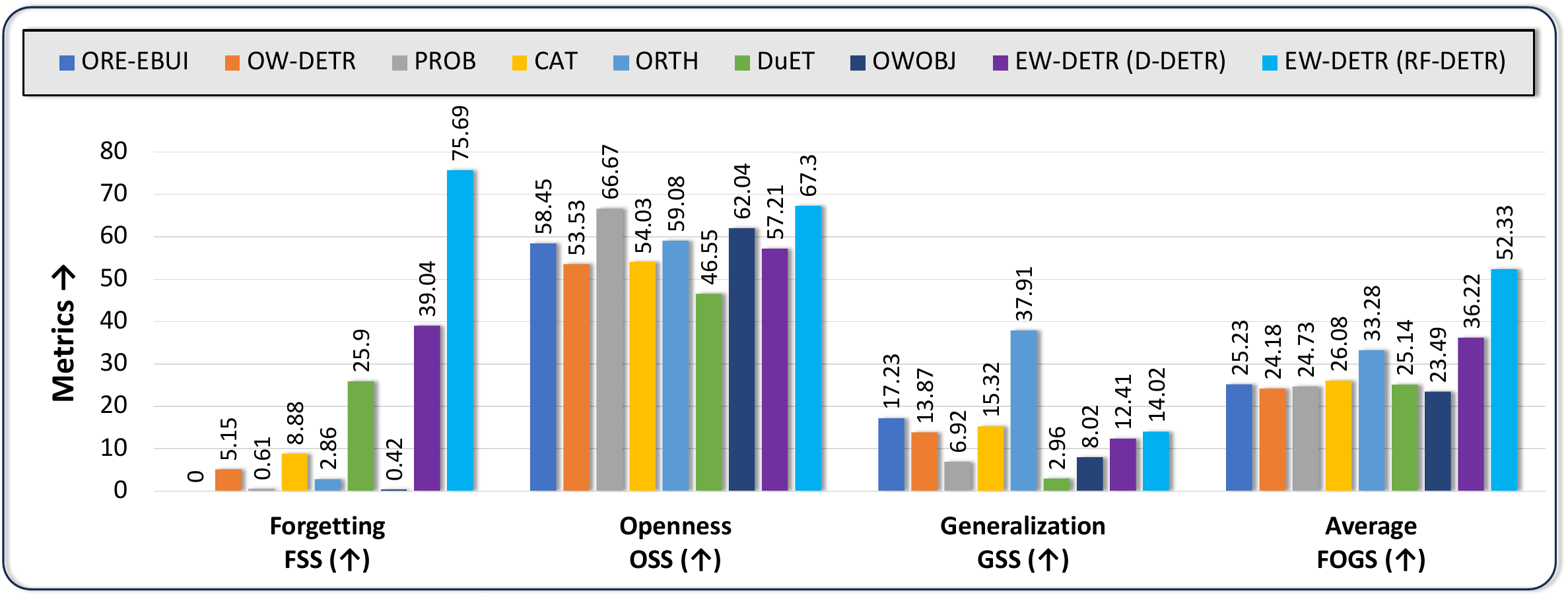}}
\vspace{-3mm}
\caption{Comprehensive comparison of various methods across all three dimensions of EWOD: Forgetting, Openness, and Generalisation, quantified by FSS, OSS, and GSS, respectively, averaged across all EWOD experiments.}
\label{fig:comparison_with_all}
\end{center}
\vspace{-8mm}
\end{figure*}

Figure~\ref{fig:comparison_with_all} presents a holistic comparison of all evaluated methods across the three critical dimensions of EWOD. EW-DETR (RF-DETR) demonstrates superior performance and achieves an average FOGS of \textbf{52.33}, which represents a \textbf{57.24\%} improvement over the next-best method ORTH~\cite{ORTH} (FOGS: 33.28). Notably, EW-DETR (RF-DETR) achieves the highest Forgetting Sub-Score (FSS: \textbf{75.69}), indicating exceptional retention of previously learned classes without exemplar replay. While PROB~\cite{PROB} demonstrates strong Openness performance (OSS: 66.67), it suffers from severe catastrophic forgetting (FSS: 0.61), highlighting the fundamental trade-off that exemplar-free methods must navigate. EW-DETR successfully balances all three dimensions, achieving competitive OSS (\textbf{67.3}) while maintaining stability (FSS: \textbf{75.69}) and reasonable generalisation (GSS: 14.02). In contrast, all OWOD methods~\cite{ORE, OW-DETR, PROB, CAT, ORTH, OWOBJ} exhibit near-zero forgetting scores, demonstrating their inability to operate under exemplar-free constraints, while DuET~\cite{DuET}, despite being an exemplar-free method, achieves moderate FSS (25.9) in comparison to EW-DETR (75.69). 

\noindent Table~\ref{tab:E1_WI_AOSE} provides a detailed breakdown of the unknown class confusion metrics: Wilderness Impact (WI)~\cite{WI} and Absolute Open-Set Error (A-OSE)~\cite{A-OSE}, that contribute to the overall Openness Sub-Score (OSS). As shown in Table~\ref{tab:E1_WI_AOSE}, EW-DETR (RF-DETR) achieves the lowest WI values across both tasks (\textbf{0.012} in $\mathcal{T}_1$ and \textbf{0.0038} in $\mathcal{T}_2$), indicating minimal precision degradation when unknown objects are present.  Moreover, it attains high Unknown Recall values (\textbf{77.35} in $\mathcal{T}_1$ and \textbf{78.23} in $\mathcal{T}_2$), demonstrating its effectiveness in identifying unknown objects without auxiliary supervision, while attaining competitive A-OSE values in comparison to other methods.

\subsection{Qualitative Visualizations}
\label{sec:qual_visuals_supp}
\begin{figure*}[ht]
\begin{center}
\centerline{\includegraphics[width=\textwidth]{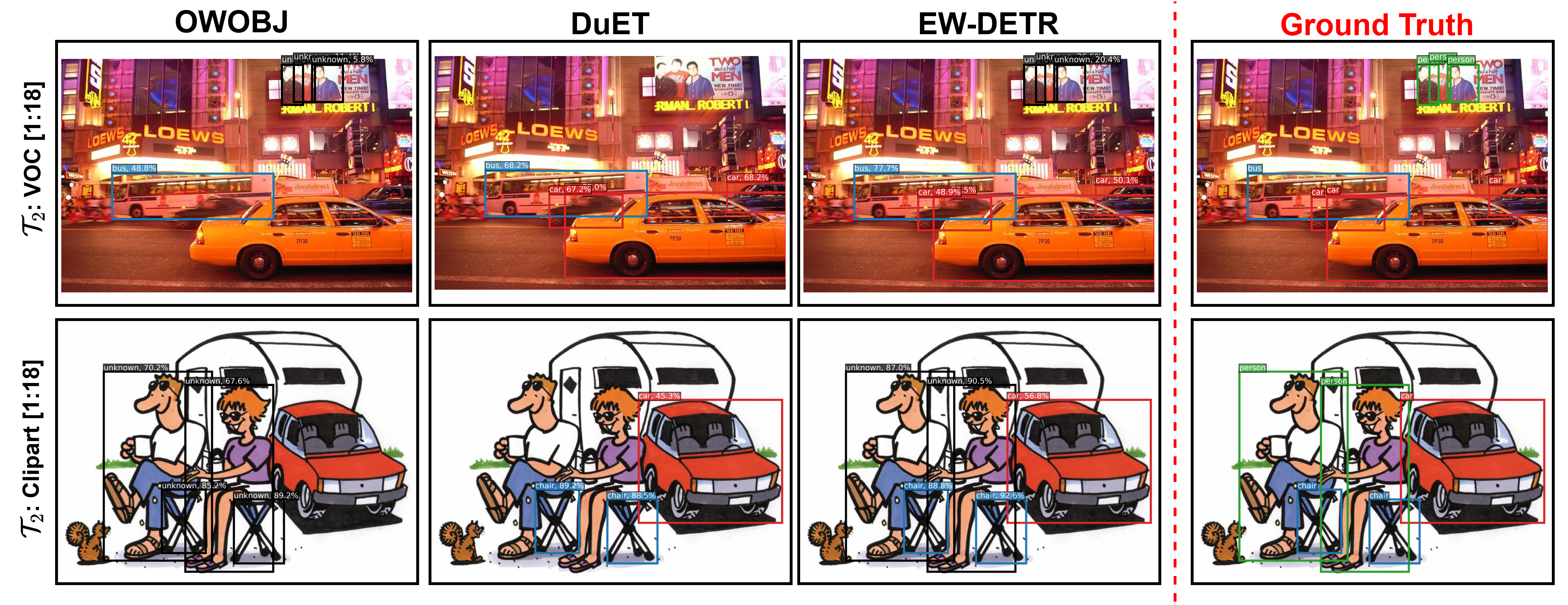}}
\vspace{-3mm}
\caption{Qualitative comparison of EW-DETR with other methods on: VOC [1:10] \(\to\) Clipart [11:18]. Best viewed in colour with zoom.}
\label{fig:qual_results}
\end{center}
\vspace{-8mm}
\end{figure*}

Figure~\ref{fig:qual_results} presents qualitative comparisons on the VOC [1:10] → Clipart [11:18] benchmark, revealing distinct failure modes of each approach. OWOBJ successfully detects unknown objects (black bounding boxes) but suffers from severe catastrophic forgetting, missing several previously learned instances, confirming its reliance on exemplar replay. DuET, operating under closed-world assumptions, absorbs unknown objects into the background. In contrast, EW-DETR correctly identifies unknown objects while simultaneously maintaining accurate detection of all previously learned instances across both domains, demonstrating an effective balance between open-set robustness, catastrophic forgetting mitigation, and cross-domain generalisation, which validates the quantitative findings in \texttt{Table 1 (main)}.

\section{Takeaways and Future Directions}
\label{sec:lim_future_works}
EW-DETR demonstrates that DETR-based detectors can effectively operate in evolving-world settings through parameter-efficient incremental adaptation, achieving strong retention (FSS: 75.69) and robust open-set detection (OSS: 67.3) without storing any previous data. However, domain generalisation (GSS: 14.02) remains modest, indicating room for improvement in cross-domain transfer under exemplar-free constraints. This work establishes EWOD as a challenging yet practical paradigm that bridges continual learning, domain adaptation, and open-world recognition. We hope our framework and comprehensive evaluation protocol inspire further research in this direction.

\clearpage
\clearpage


\end{document}